\documentclass{article}

\usepackage{arxiv}
\usepackage[square,numbers]{natbib}

\usepackage[utf8]{inputenc} %
\usepackage[T1]{fontenc}    %
\usepackage[hidelinks]{hyperref}       %
\usepackage{url}            %
\usepackage{booktabs}       %
\usepackage{amsfonts}       %
\usepackage{nicefrac}       %
\usepackage{microtype}      %
\usepackage{xcolor}         %
\usepackage{graphicx}
\usepackage{tikz}

\usepackage[acronym]{glossaries}

\newacronym{ai}{AI}{Artificial Intelligence}
\newacronym{dl}{DL}{Deep Learning}
\newacronym{dnn}{DNN}{Deep Neural Network}
\newacronym{diffae}{DiffAE}{Diffusion Auto-Encoder}
\newacronym{lrp}{LRP}{Layer-wise Relevance Propagation}
\newacronym{lsb}{LSB}{least significant bit}
\newacronym{xai}{XAI}{eXplainable Artificial Intelligence}
\newacronym{crp}{CRP}{Concept Relevance Propagation}
\newacronym{amax}{ActMax}{Activation Maximization}
\newacronym{rmax}{RelMax}{Relevance Maximization}
\newacronym{auc}{AUC}{Area Under Curve}
\newacronym{llm}{LLM}{Large Language Models}
\newacronym{aoc}{AOC}{Area Over Curve}
\newacronym{conv}{Conv}{convolutional}
\newacronym{svm}{SVM}{Support Vector Machine}
\newacronym{roi}{ROI}{Region of Interest}
\newacronym{lcrp}{L-CRP}{CRP for Localization Models}
\newacronym{rrr}{RRR}{Right for the Right Reason}
\newacronym{cdep}{CDEP}{Contextual Decomposition Explanation Penalization}
\newacronym{clarc}{ClArC}{Class Artifact Compensation}
\newacronym{aclarc}{\mbox{A-ClArC}}{Augmentive ClArC}
\newacronym{pclarc}{\mbox{P-ClArC}}{Projective ClArC}
\newacronym{rrclarc}{RR-ClArC}{Right Reason ClArC}
\newacronym{ml}{ML}{Machine Learning}
\newacronym{cse}{CSE}{complete skin examination}
\newacronym{cav}{CAV}{Concept Activation Vector}
\newacronym{tcav}{TCAV}{Testing with CAV}
\newacronym{spray}{SpRAy}{Spectral Relevance Analysis}
\newacronym{iterrev}{IterRev}{Iteratively Revealing and Revising Spurious Model Behavior}
\newacronym{r2r}{R2R}{Reveal to Revise}
\newacronym{xil}{XIL}{eXplanatory Interactive Learning}
\newacronym{sem}{SEM}{Standard Error of the Mean}
\newacronym{se}{SE}{Standard Error}
\newacronym{sae}{SAE}{Sparse Autoencoder}
\newacronym{vit}{ViT}{Vision Transformer}

\newcommand*\circledblacksmall[1]{\tikz[baseline=(char.base)]{
            \node[shape=circle,draw,inner sep=1pt,fill=blue!40,color=blue!40] (char) {\color{white}\footnotesize{#1}};}}

\newcommand*\circledredsmall[1]{\tikz[baseline=(char.base)]{
            \node[shape=circle,draw,inner sep=1pt,fill=red!60,color=red!60] (char) {\color{white}\small{#1}};}}

\usepackage{cleveref}
\crefname{figure}{Fig.}{Figs.}
\crefname{table}{Tab.}{Tabs.}
\crefname{equation}{Eq.}{Equations}
\crefname{section}{Sec.}{Sections}
\crefname{appendix}{App.}{Appendix}

\title{From What to How: Attributing CLIP's Latent Components Reveals Unexpected Semantic Reliance}

\author{%
Maximilian Dreyer$^{1}$ \quad Lorenz Hufe$^{1}$ \quad Jim Berend$^1$ \\ \textbf{Thomas Wiegand$^{1,2,3}$ \quad Sebastian Lapuschkin$^{1,4}$ \quad Wojciech Samek$^{1,2,3}$}\\
$^1$Fraunhofer Heinrich Hertz Institute \quad $^2$Technische Universität Berlin \\
$^3$BIFOLD -- Berlin Institute for the Foundations of Learning and Data\\
$^4$Centre of eXplainable Artificial Intelligence, Technological University Dublin \\
\texttt{\{wojciech.samek,sebastian.lapuschkin\}@hhi.fraunhofer.de}\\
}

\date{}

\begin{document}

\maketitle

\begin{abstract}
Transformer-based CLIP models are widely used for text-image probing and feature extraction, making it relevant to understand the internal mechanisms behind their predictions. While recent works show that Sparse Autoencoders (SAEs) yield interpretable latent components, they focus on \emph{what} these encode and miss \emph{how} they drive predictions. We introduce a scalable framework that reveals what latent components activate for, how they align with expected semantics, and how important they are to predictions. To achieve this, we adapt attribution patching for instance-wise component attributions in CLIP and highlight key faithfulness limitations of the widely used Logit Lens technique. By combining attributions with semantic alignment scores, we can automatically uncover reliance on components that encode semantically unexpected or spurious concepts. Applied across multiple CLIP variants, our method uncovers hundreds of surprising components linked to polysemous words, compound nouns, visual typography and dataset artifacts. While text embeddings remain prone to semantic ambiguity, they are more robust to spurious correlations compared to linear classifiers trained on image embeddings. A case study on skin lesion detection highlights how such classifiers can amplify hidden shortcuts, underscoring the need for holistic, mechanistic interpretability.
We provide code at \url{https://github.com/maxdreyer/attributing-clip}.
\end{abstract}

\section{Introduction}

Vision-language models such as CLIP~\cite{radford2021learning} are trained in semi-supervised fashion on paired images and text to learn a wide range of visual concepts without requiring explicit human annotation. 
These models are widely adopted as feature encoders in larger vision-language systems for tasks such as retrieval, captioning, and reasoning~\cite{li2023blip,zhu2024minigpt}.
While semi-supervised training is scalable and enables strong generalization across diverse domains, it also raises fundamental questions: What kinds of patterns do CLIP models rely on? And how do these align -- or misalign -- with human expectations?

Prior work has focused on explaining CLIP’s representations via human-expected concepts, relying on predefined concept sets~\cite{bhalla2024interpreting,yun2023do}. However, such approaches tend to overlook features that emerge from biases in image-text data, including spurious correlations~\cite{wang2024a} and ambiguous linguistic cues~\cite{kumar2024vision}. As a result, key aspects of model behavior may remain hidden.

An alternative line of research investigates internal components of neural networks, such as neurons or latent dimensions. These methods aim to provide an explanation of the full spectrum of learned semantic signals that a model perceives. For convolutional neural networks, recent frameworks such as SemanticLens~\cite{dreyer2025mechanistic} and WWW~\cite{ahn2024unified} have enabled such analyses by measuring what components encode (by providing textual descriptions) and how they influence model outputs via latent attribution scores. However, these techniques are not directly applicable to CLIP’s \gls{vit}-based architecture due to its lack of interpretable intermediate representations and the absence of attribution methods in latent space.

While the Logit Lens~\cite{belrose2023eliciting} approach for probing component$\leftrightarrow$output relations in language models offers a \emph{global} alignment measure~\cite{bloom2024understanding}, it does not capture \emph{instance-wise} attribution or interaction effects~\cite{gandelsman2025interpreting}. Alternatively, attribution patching~\cite{nanda2023attribution}, which is related to Input$\times$Gradient~\cite{shrikumar2017learning} attributions in latent space, represents an efficient alternative for attributing weights or components in language models relevant for a specific behavior.
\begin{figure}[t]
    \centering
    \includegraphics[width=0.99\linewidth]{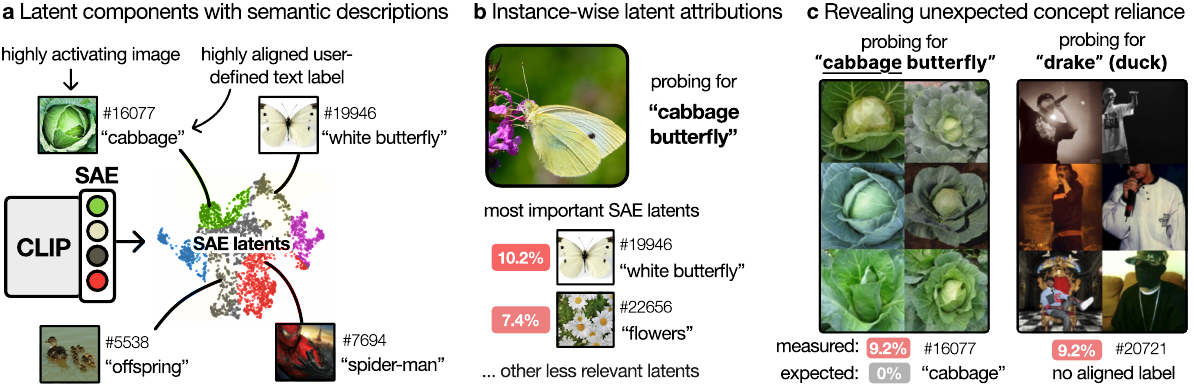}
    \caption{A framework for interpreting CLIP via latent attributions.
a) Sparse autoencoders (SAEs) extract a diverse set of interpretable latent components from CLIP representations. Each component is assigned a textual label from an expected set of concepts.
b) Instance-wise attribution scores quantify each component’s contribution to model predictions.
c) Attribution reveals unexpected model behavior, 
including components that are unusually predictive relative to a baseline expectation (e.g., on a test set) or lack strong semantic alignment to expected textual labels.}
    \label{fig:introduction}
\end{figure}

In this work, we adapt attribution patching for text-image probing with CLIP, enabling a more faithful quantification of how latent components contribute to outputs compared to Logit Lens, as illustrated in \cref{fig:introduction}. We further build on recent advances in sparse dictionary learning and employ \glspl{sae}~\cite{huben2024sparse,bricken2023towards} to extract interpretable components from CLIP’s latent representations.  By combining SAEs with attributions, we provide the first framework for scalable component-level analysis of CLIP models that explains both {input}$\leftrightarrow${component} and {component}$\leftrightarrow${output} relations.

Our approach, applied to the final embeddings after the last transformer block, reveals the relevance of hundreds of unexpected or unintended features, often arising from polysemous words, compound nouns, visual typography, or spurious correlations. We benchmark various CLIP architectures and show that even larger models exhibit such issues, suggesting practical limitations in how textual inputs are encoded. Finally, we apply our framework to the domain of skin lesion classification, where we identify and correct the reliance on background features, measurably improving robustness.

\textbf{Contributions:}
The contributions of this work are summarized as follows:
\begin{itemize}
\item \textbf{Instance-wise component attribution:} We adapt attribution patching to CLIP, enabling instance-specific attribution of latent components. We draw an intrinsic connection to the Logit Lens technique and highlight its limitations for achieving locally faithful explanations.

\item \textbf{Diverse and interpretable components with SAEs:} Through sparse autoencoding, we discover a rich set of interpretable components within CLIP models. Notably, highly relevant latents tend to be more interpretable than weak latents.

\item \textbf{A holistic framework for revealing failure cases:} We combine latent attributions with semantic descriptions of highly activating samples, allowing to analyze \emph{what} and \emph{how} concepts are used. This enables to probe for outlier behaviors by identifying relevant concepts that deviate from expected semantics or expected importance patterns.
\item \textbf{Systematic analysis of text-embedding failures:} We identify persistent issues in text embeddings across multiple CLIP variants, stemming from ambiguous textual alignment.
However, compared to training linear classifiers on visual embeddings, text-based probing shows greater robustness against dataset-specific spurious correlations.

\item \textbf{Case study for downstream applications:} In a skin lesion classification task, we demonstrate that naively training linear classifiers on CLIP embeddings risks amplifying hidden spurious correlations, thereby reducing robustness in application.
\end{itemize}

\section{Related work}

\textbf{Understanding CLIP representations:}
A line of work on interpreting CLIP and other vision-language models focuses on decomposing latent embeddings into a set of predefined, human-interpretable concepts~\cite{bhalla2024interpreting,yun2023do}. These methods generally evaluate alignment to known textual labels but struggle to capture the full diversity of learned features, particularly unexpected or spurious ones.

For a more comprehensive understanding, component-level understanding is essential. 
To this end, studies have begun to examine attention heads and MLP neurons~\cite{gandelsman2024interpreting,gandelsman2025interpreting,dorszewski2025colors}. Alternatively, visual embeddings in CLIP's last layer can be decomposed into a diverse set of interpretable components using \glspl{sae}~\cite{pach2025sparse,rao2024discover,joseph2025steering}, which have been originally proposed for language models~\cite{huben2024sparse,bricken2023towards}. 
Rao et al.~\cite{rao2024discover} train linear classifiers directly on \glspl{sae} activations for downstream-tasks, but, as such, can no longer explain the original CLIP model.
Other works based on \glspl{sae} \cite{lim2025sparse,thasarathan2025universal} successfully explain \emph{what} components encode and represent, yet overlook \emph{how} they influence specific predictions.

\textbf{Holistic interpretability frameworks:}
Comprehensive interpretability frameworks aim to analyze both {input}$\leftrightarrow${component} and {component}$\leftrightarrow${output} relations.
CRP~\cite{achtibat2023attribution} and CRAFT~\cite{fel2023craft} map neurons to highly activating examples and compute relevance scores for predictions.  WWW~\cite{ahn2024unified} and  SemanticLens~\cite{dreyer2025mechanistic} further assign textual labels to components. While these approaches allow for a comprehensive understanding effective for debugging and scientific discovery, they are typically limited to individual filters or neurons in convolutional networks. 
\Glspl{sae} enable the extraction of interpretable components for CLIP, but how to compute attributions remains an open challenge.

\textbf{Latent attributions:}
For convolutional networks and image data, feature attribution methods such as Input$\times$Gradient~\cite{shrikumar2017learning}, GradCAM~\cite{selvaraju2017grad}, and LRP~\cite{bach2015pixel} have been developed to identify input pixels relevant to a model's prediction. Recent work on vision models has extended input attributions to latent representations, enabling both local~\cite{achtibat2023attribution,fel2023holistic} and global~\cite{dreyer2024understanding} analysis of latent components.

For interpreting transformer-based language models, Logit Lens~\cite{belrose2023eliciting} has emerged as a popular tool by projecting intermediate activations directly to output logits~\cite{bloom2024understanding}. While this method provides a global measure for output alignment, it is limited in its ability to reveal which specific latents contribute to particular outputs. 
Alternatively, activation patching has been proposed as a causal intervention technique to identify neurons and weights responsible for specific behaviors~\cite{vig2020investigating}. As this method requires multiple forward passes, attribution patching based on output gradients has been proposed as a more efficient alternative~\cite{nanda2023attribution}, also applied for \gls{sae} components recently~\cite{syed2024attribution,marks2025sparse}.

While there is growing interest in interpreting large vision-language models, few approaches offer holistic and scalable tools that operate at the component level. Attribution scores for \gls{vit}s and CLIP have remained largely unexplored, and existing tools typically fall short in revealing how exactly CLIP relies on its components.
Our work bridges this gap by combining SAEs with a novel adaptation of attribution patching for CLIP. This enables systematic identification of the components most responsible for specific predictions, even when they encode unexpected or spurious concepts. Unlike previous work, our framework is designed to support the full pipeline of discovering, labeling,
and attributing latent components in CLIP models.
\section{Prelimenaries}
The following section introduces \glspl{sae}, textual labeling of components and the Logit Lens technique.
\subsection{Sparse autoencoders for extracting interpretable components}
One major challenge in transformer interpretability lies in the often non-interpretable and polysemantic nature of individual neurons \cite{elhage2022toymodelssuperposition}. Motivated by the sparse feature hypothesis \cite{olshausen1997sparse}, recent advancements in sparse dictionary learning \cite{huben2024sparse,bricken2023towards} have shown that \glspl{sae} can effectively identify interpretable monosemantic directions in representations. 
For a latent representation $\mathbf{x} \in \mathbb{R}^{d}$ of dimension $d$ in a hidden layer, the \gls{sae} computes a decomposition as follows:
\begin{equation}
    \label{eq:background:sae}
{\mathbf{x}} 
= \hat{\mathbf{x}} (\mathbf x) + \boldsymbol{\varepsilon}(\mathbf{x})
= \sum_{i=1}^{d_{\text{SAE}}} a_i(\mathbf{x})\mathbf{v}_i
+ \mathbf{b} + \boldsymbol{\varepsilon}(\mathbf{x}) ~.
\end{equation}
This formulation reconstructs the representation $\mathbf{x}$ through a sparse combination of $d_\text{SAE}$ feature vectors $\mathbf{v}_i \in \mathbb{R}^{d}$, with activations $a_i(\mathbf{x}) \in \mathbb{R}$, bias term $\mathbf{b} \in \mathbb{R}^{d}$, and an error term $\boldsymbol{\varepsilon}(\mathbf{x}) = \mathbf{x} - \hat{\mathbf{x}}(\mathbf{x})$.

The SAE is post-hoc trained and optimized to reduce the reconstruction error $\boldsymbol{\varepsilon}(\mathbf{x})$ and constraint to enforce activation sparsity, e.g., via $L_1$-norm regularization. Recently, top-${k}$ activation functions have been introduced, where activations $a_i(\mathbf{x})$ are set to zero if they do not belong to the set of the  ${k}$  largest activations~\cite{gao2025scaling}, a simple but effective measure to enforce sparsity.
Througout experiments in \cref{sec:exp}, we use top-${k}$ \glspl{sae}, and set $\|\mathbf{v}_i \|=1$ without loss of generality.

\subsection{Understanding what semantics components encode}
\begin{figure}[t]
    \centering    \includegraphics[width=0.99\linewidth]{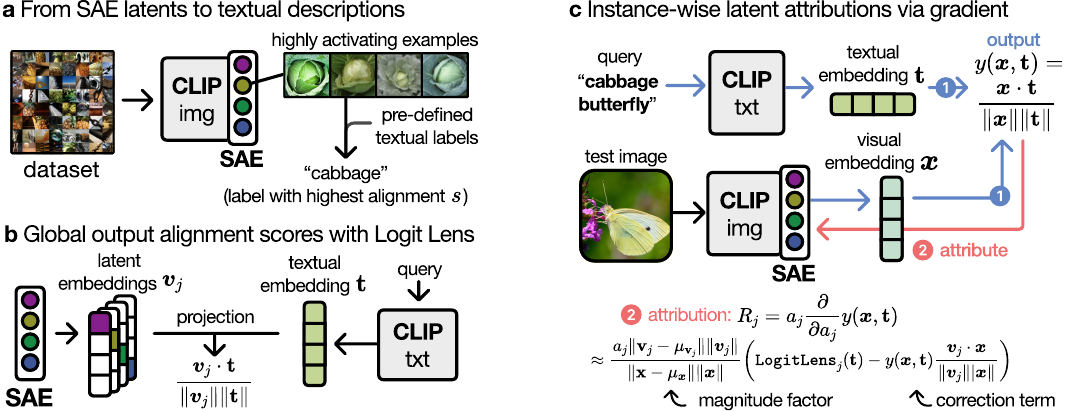}
    \caption{Methodological overview:
    a) For each \gls{sae} component, we collect its most activating samples and assign a textual label based on semantic alignment.
    b) Logit Lens computes a global alignment score by projecting each latent embedding onto a textual embedding.
    c) Instance-wise attributions are derived by \protect\circledblacksmall{1} performing a forward pass to obtain predictions and \protect\circledredsmall{2} backpropagating gradients to estimate each component’s contribution. In contrast to Logit Lens, this -- applied after the last transformer block -- accounts for component magnitude (including activation) and adds a local correction term.
    }
    \label{fig:method:method}
\end{figure}
The activation of a component signals the presence of a specific visual input pattern, commonly referred to as ``concept''. To communicate this concept, the top-$k$ most activating samples from a dataset are usually retrieved, as illustrated in \cref{fig:method:method}a. For textual labeling of a component $j$, we follow SemanticLens, which uses Mobile-CLIP-S2~\cite{vasu2024mobileclip} to first compute the average visual CLIP embedding $\bar{\boldsymbol{x}}_j$ of these examples. Note, that $\boldsymbol{x}$ represents the last transformer layer's class token embedding $\mathbf{x}_\text{cls}$ after a LayerNorm and projection operation, as is standard in CLIP.
We then compute the alignment between average embedding $\bar{\boldsymbol{x}}_j$ and each text embedding $\mathbf{t} \in \mathcal{T}$ of predefined labels, and further subtract the alignment with an empty prompt $\mathbf{t}_\text{empty}$:
\begin{equation}
\label{eq:background:labeling}
    s_j(\mathbf{t})
    = \frac{\bar{\boldsymbol{x}}_j \cdot \mathbf{t}}{\|\bar{\boldsymbol{x}}_j\| \|\mathbf{t}\|} - \frac{\bar{\boldsymbol{x}}_j \cdot \mathbf{t}_\text{empty}}{\|\bar{\boldsymbol{x}}_j\| \|\mathbf{t}_\text{empty}\|} \,,
\end{equation}
where $s_j(\mathbf{t})$ is high when the label $\mathbf{t}$ semantically matches the component's highly activating samples, and negative if the empty prompt aligns better. The component label is assigned as $\text{argmax}_{\mathbf{t} \in \mathcal{T}} s_j(\mathbf{t})$.

\subsection{Understanding component-output alignment with Logit Lens}
Logit Lens is a technique for estimating how individual components align with specific outputs by isolating their contribution to the model’s prediction. The core assumption is that a single component is solely responsible for the embedding $\mathbf{x}$, which is then propagated to produce output logits. This yields a global alignment score that reflects how well a given representation aligns with the final output, offering insight into how the model might use that component.

In the context of CLIP and \glspl{sae}, this procedure involves projecting the intermediate representation $\mathbf{v}_j$ of component $j$ onto a textual embedding $\mathbf{t}$. This corresponds to $a_i = \delta_{ij}$ and $\mathbf{b} = \boldsymbol{\varepsilon}(x) \overset{!}{=} 0$ in \cref{eq:background:sae}, where $\delta_{ij}$ denotes the Kronecker delta.
After applying the downstream LayerNorm and projection layer, we obtain the embedding ${\boldsymbol{v}}_k$, and compute the Logit Lens alignment score as:

\begin{equation}
\label{eq:preliminaries:logitlens}
    \texttt{LogitLens}_j(\mathbf{t}) = \frac{{\boldsymbol{v}}_j}{\| {\boldsymbol{v}}_j\|} \cdot \frac{\mathbf{t}}{\| \mathbf{t} \|}~.
\end{equation}

Notably, Logit Lens assumes activation of a \emph{single} latent, ignoring magnitude and interactions with other components, providing only a global approximation of a component’s relevance to an output.
\section{Method}
\label{sec:methods}
We present a holistic framework for analyzing the latent representations of CLIP models.
Building on \glspl{sae}, we extract interpretable components and compute their semantic alignment to expected concepts (e.g., textual labels) via \cref{eq:background:labeling}.
Our core methodological contributions are: (1) the computation of instance-wise attribution scores for individual latent components, and (2) the automatic detection of outlier concept reliance.

CLIP’s image encoder consists of transformer blocks, each producing token-wise embeddings of shape $d_\text{token} \times d_\text{pre}$, where each token (including spatial and class tokens) has an embedding $\mathbf{x} \in \mathbb{R}^{d_\text{pre}}$.
After the final transformer block, the output is processed through:
1) chosing the class token embedding, 2) LayerNorm with optional scaling and bias $\boldsymbol{\gamma},\boldsymbol{\beta}\in\mathbb{R}^{d_\text{pre}}$,  and 3) a projection through matrix $W_\text{proj}\in \mathbb{R}^{d_\text{pre}\times d_\text{post}}$ to dimension $d_\text{post}$.
The model prediction $y$ for an image-text pair is computed via cosine similarity between the projected image embedding ${\boldsymbol{x}}$ and a textual embedding $\mathbf{t}$:
\begin{equation}
    y ({\boldsymbol{x}}, \mathbf{t})= \frac{{\boldsymbol{x}}}{\| {\boldsymbol{x}}\|} \cdot \frac{\mathbf{t}}{\| \mathbf{t}\|} 
    , \quad \text{where} \quad 
    \boldsymbol{x}= \text{LayerNorm}_{\boldsymbol{\gamma},\boldsymbol{\beta}}\left[\mathbf{x}\right]W_\text{proj}~.
\end{equation}

While our framework can be applied at any transformer layer, we focus on the final block’s class token embedding for clarity, which is also used directly for prediction.

\subsection{Instance-wise attribution of components}
\label{sec:methods:attributions}

In language models, attribution patching quantifies how a model’s output changes when a neuron’s activation is replaced with a reference value \( a_\text{ref} \), using a first-order Taylor approximation. We extend this approach to CLIP by estimating the effect of zeroing out individual latent activations. Specifically, we analyze the output \( y(\boldsymbol{x}, \mathbf{t}) \) for \( a_\text{ref} = 0 \), yielding component-wise attributions \( R_j(\boldsymbol{x}, \mathbf{t}) \) defined as
\begin{align}
\label{eq:methods:relevance}
    R_j(\boldsymbol{x}, \mathbf{t})
    &= y(\boldsymbol{x}, \mathbf{t}) - y(\boldsymbol{x}, \mathbf{t})\big|_{a_j = 0}
    \approx a_j\frac{\partial y(\boldsymbol{x}, \mathbf{t})}{\partial a_j}~,
\end{align}
resembling Input$\times$Gradient~\cite{shrikumar2017learning} attributions in the latent space (i.e., ``Activation$\times$Gradient''), which showed faithful results for convolutional networks before~\cite{dreyer2024understanding}. 
As derived in \cref{app:sec:latent_attributions}, when assuming that the LayerNorm bias $\boldsymbol{\beta}$ is small, 
we receive the approximation for $R_j(\boldsymbol{x}, \mathbf{t})$ of
\begin{equation}
\label{eq:methods:approximation}
    a_j\frac{\partial y(\boldsymbol{x}, \mathbf{t})}{\partial a_j} \overset{\boldsymbol{\beta} 
    \rightarrow 0}{=}  
    a_j \frac{\|\boldsymbol{v}_j\|}{\|\boldsymbol{x}\|}\frac{\|\mathbf{v}_j - \mu_{\mathbf{v}_j}\|}{\|\mathbf{x} - \mu_\mathbf{x}\|}  
    \left( \texttt{LogitLens}_j(\mathbf{t}) - y(\boldsymbol{x}, \mathbf{t}) \frac{\boldsymbol{v}_j}{\|\boldsymbol{v}_j\|} \cdot \frac{\boldsymbol{x}}{\|\boldsymbol{x}\|} \right)~,
\end{equation}
where $\mu_{\mathbf{x}/\mathbf{v}_j}$ denotes the mean over the elements of $\mathbf{x}$ or $\mathbf{v}_j$, respectively.
This highlights the distinction from the Logit Lens method: an additional scaling factor that captures the strength of the latent component (including its activation), as well as a correction term. The correction term reduces the relevance score when the output is already high, and the projected embedding $\boldsymbol{v}_j$ is highly aligned with embedding $\boldsymbol{x}$.

\subsection{Revealing unexpected concept reliance}
\label{sec:methods:attribution_analysis}
We propose two complementary strategies to identify concepts that models unexpectedly rely on by combining latent relevance scores with either user-defined or statistically derived expectations.

\textbf{a) Hidden concepts: high relevance, low alignment:}
This approach identifies model components that are highly relevant to a prediction but poorly aligned with known or expected semantics. Specifically, for each component $j$, we compute a relevance score $R_j(\boldsymbol{x}, \mathbf{t})$, and a textual alignment score $s_j(\mathbf{t})$ (see \cref{eq:background:labeling}) to predefined label embeddings $\mathbf{t}\in \mathcal{T}$.
We then select components that satisfy:
\begin{equation}
R_j(\boldsymbol{x}, \mathbf{t}) \geq \tau_\text{rel}, \quad s_j(\mathbf{t}) \leq \tau_\text{align}
\end{equation}
where \( \tau_\text{rel} \) and \( \tau_\text{align} \) are thresholds for relevance and alignment respectively.
Thresholding high relevance scores effectively filters out noisy or irrelevant components, isolating those that meaningfully contribute to predictions. These identified components often unveil novel and unexpected features. Alternatively, activations $a_j$ can serve as an output-agnostic proxy for relevance (see \cref{eq:methods:approximation}).

\textbf{b) Failure cases: higher relevance than expected:}
We define an expected relevance distribution for each component based on a reference set (e.g., training images with known labels). For component $j$, we collect relevance scores across this reference set, calculating the sample mean $\mu_j$ and standard deviation $\hat{\sigma}_j$. Given a new input with embedding \( \boldsymbol{x}_\text{test} \), we compute its relevance $R_j(\boldsymbol{x}_\text{test}, \mathbf{t})$, and derive the $z$-score to determine whether this value deviates significantly from the expected distribution. The $z$-score, which serves as a standard method to detect outliers~\cite{iglewicz1993volume}, quantifies the deviation in terms of standard deviations from the mean:
\begin{equation}
\label{eq:methods:zscore}
     z = \frac{R_j(\boldsymbol{x}_\text{test}, \mathbf{t}) - \mu_j}{\hat{\sigma}_j}.
\end{equation}
Components yielding a \( z \)-score above $3.0$ are commonly flagged as potential failure points, indicating that the model relies on them more than expected and should be inspected.
This approach assumes approximate normality of relevance scores, or sufficient \( n \) for the Central Limit Theorem to hold.

\label{sec:exp:faithfulness}
\begin{figure}[t]
    \centering
    \includegraphics[width=0.99\linewidth]{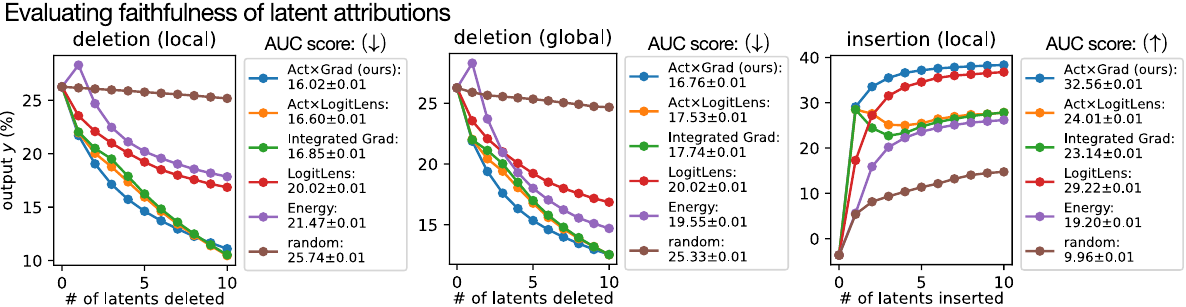}
    \caption{Faithfulness evaluation of latent attributions on the ImageNet test set. We measure output scores while performing latent deletion (setting activations to zero) with instance-wise attributions (\emph{left}) and average attributions on a subset (\emph{middle}), and while performing latent insertion (\emph{right}).}
    \label{fig:experiments:faithfulness}
\end{figure}

\section{Experiments}
\label{sec:exp}
We design our experiments to answer the following key research questions:
\begin{enumerate}
    \item[\textbf{Q1}] How faithfully do attributions reflect a component’s actual influence on CLIP predictions?
    \item[\textbf{Q2}] How effective are SAEs for extracting human-interpretable components in CLIP, and how many unexpected concepts exist?
    \item[\textbf{Q3}]How robust is text-image probing in CLIP and what are typical failure cases?
    \item[\textbf{Q4}] Can we detect spurious behavior in a CLIP-based classifier trained for melanoma detection?
\end{enumerate}
\textbf{Experimental settings:}
We train top-$k$ \glspl{sae} with $k=64$ and 30,000 components on the class token embeddings of the last transformer block, using the ImageNet-1k~\cite{krizhevsky2012imagenet} train set. Experiments cover publicly available \gls{vit} variants: B/32, B/16, L/14, and H/14~\cite{ilharco_gabriel_2021_5143773}. To assess semantic alignment of SAE latents, we evaluate alignment against ImageNet-21k class names~\cite{ridnik2021imagenetk} using CLIP-Mobile-S2. For the medical domain, we apply our method to WhyLesion-CLIP~\cite{yang2024a} and train SAEs on ISIC skin lesion datasets~\cite{tschandl2018ham10000,codella2018skin,hernandez2024bcn20000,rotemberg2021patient,international2024slice}. Additional implementation details are provided in \cref{app:sec:experimental_settings}.

\subsection{Evaluating attribution faithfulness (Q1)}

We assess whether component attribution scores reflect the true influence of individual latents on model predictions by performing deletion and insertion experiments. In the deletion setting, we ablate latents by setting their activations to zero, starting from the most relevant components. In the insertion setting, we begin with all latent activations ablated and sequentially reintroduce them in order of estimated relevance. We perform both local (instance-wise relevance) and global (mean relevance over a reference set) deletion experiments. All evaluations are conducted on the ImageNet test set for 100 classes, with 50 samples per class and prompting the CLIP \gls{vit}-L/14 model with the class name as text input. We compare six methods: our proposed Act$\times$Grad score as in \cref{eq:methods:relevance}, the product of activation and Logit Lens score, Logit Lens alone, Energy~\cite{thasarathan2025universal} (given as activation $a_j$ times the norm of $\mathbf{v}_j$, see \cref{eq:background:sae}), Integrated Gradients~\cite{marks2025sparse}, and a random ordering of active latents. 

As shown in \cref{fig:experiments:faithfulness}, ~Act$\times$Grad attributions consistently lead to the largest performance drop in deletion and the fastest recovery in insertion, indicating higher faithfulness. AUC scores with \gls{sem} across 9 subsets are reported in the legend. Activation times Logit Lens provides improvements over Logit Lens in deletion, but remains inferior to our method, especially in cases involving few interdependent latents.
Results on other CLIP models are provided in \cref{app:sec:latent_attributions}.

\subsection{Interpretability and diversity of SAE components  (Q2)}
\label{sec:exp:sae}
\begin{figure}[t]
    \centering
    \includegraphics[width=0.97\linewidth]{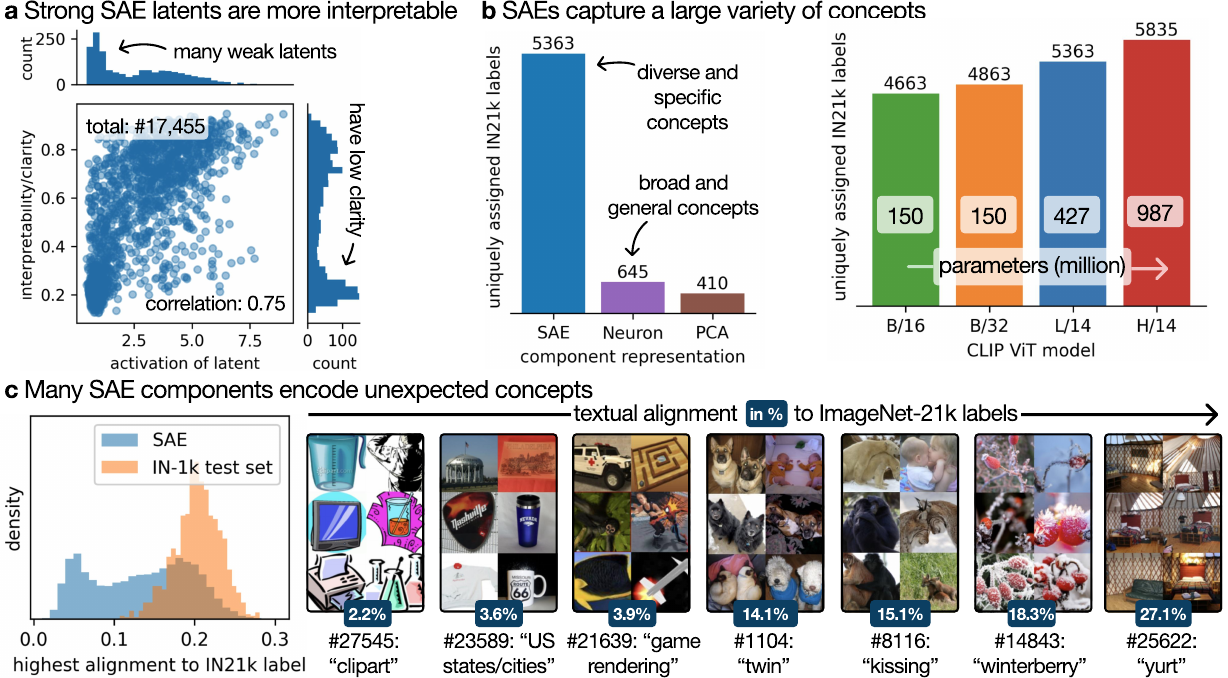}
    \caption{Analysis of components extracted by SAEs in CLIP.
    a) Most latent components have both low activation and low interpretability; activation strongly correlates with interpretability.
    b) Compared to the inherent neural basis and PCA, SAEs yield a significantly more diverse and specific set of concepts. Larger CLIP models tend to encode a broader range of semantic concepts.
    c) When probing components with expected labels, a large fraction show weak alignment (compared to alignment between ImageNet-1k class names and test set images). Unexpected concepts include game renderings, cliparts, and visually similar object pairs. More examples can be found in \cref{app:sec:sae_interpretability}.
    }
    \label{fig:experiments:sae_motivation}
\end{figure}
We study the number, interpretability, and diversity of latent components extracted from CLIP models using \glspl{sae}. For the \gls{vit}-L/14 model, we obtain 17,455 components (filtered to those with $\geq20$ non-zero activations for meaningful statistical analysis of interpretability) from an SAE trained on the final class token embeddings . To assess interpretability, we use the clarity metric introduced in~\cite{dreyer2025mechanistic}, which quantifies visual similarity among top-activating samples. As shown in \cref{fig:experiments:sae_motivation}a, the majority of components exhibit unexpectedly low clarity.
Analyzing the relationship between interpretability and activation scores on the 20 highly activating samples, we observe a strong correlation of $ 0.75$, indicating that components with higher activations (which relate to relevance via \cref{eq:methods:relevance}) tend to be more interpretable. Notably, activation frequency and activation magnitude show a smaller correlation with interpretability (see \cref{app:sec:sae_interpretability}).

To evaluate representational diversity, we compare the number of unique concepts identified across different bases: the inherent neural basis, PCA-transformed embeddings, and \gls{sae} feature space. \glspl{sae} yield a larger and more specific concept set (5,363 unique labels vs. less than $500$ for PCA). Moreover, larger CLIP models capture broader concept sets, with ViT-H/14 showing over 20\,\% more unique concepts than ViT-B/16, indicating that model capacity contributes to conceptual diversity.

Finally, probing latent components with ImageNet-21k class names using CLIP-Mobile-S2 for alignment scoring (filtered with $\tau_\text{act} = 3.0$ as in method \cref{sec:methods:attribution_analysis}a), we uncover a substantial fraction of components encoding semantically unexpected features. These include high-level visual abstractions such as ``clipart'', ``game rendering'', or ``US states'', which are not part of the labeled class taxonomy (see \cref{fig:experiments:sae_motivation}c). This highlights the presence of latent knowledge in foundation models that extends beyond standard dataset annotations. Additional examples are provided in \cref{app:sec:sae_interpretability}.

\subsection{Robustness of CLIP to semantic ambiguity and spurious correlations  (Q3)}
\label{sec:exp:robustness}
\begin{figure}[t]
    \centering
    \includegraphics[width=0.99\linewidth]{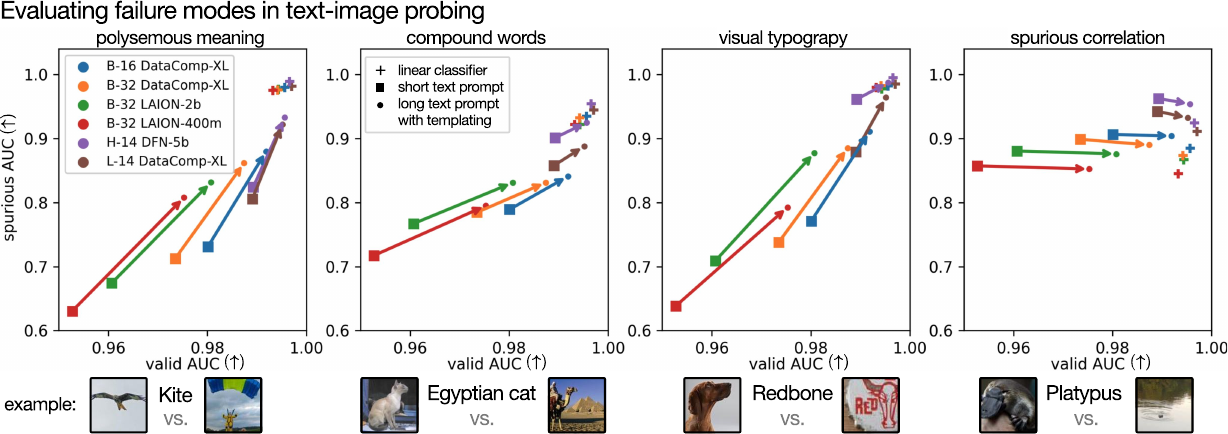}
    \caption{Analysis of failure modes in text-image probing. We evaluate CLIP’s robustness to textual ambiguities and dataset artifacts across multiple model variants. Failure cases include polysemous words, compound nouns, typography in images, and spurious correlations. For each scenario, we assess the separability of true images from found distractors (``spurious AUC'') and from unrelated ImageNet classes (``valid AUC'').
    We compare performance using both vanilla and enriched text prompts (e.g., via templating and detailed descriptions), and include linear classifiers trained on image embeddings as a baseline. Uncertainty estimates are provided in \cref{app:sec:robustness}.}
    \label{fig:experiments:benchmark}
\end{figure}
We evaluate the robustness of CLIP models to semantic ambiguities and spurious correlations across multiple \gls{vit} variants. Following the approach introduced in \cref{sec:methods:attribution_analysis}b, we identify latent components that contribute to high-confidence predictions for a given ImageNet-1k class significantly more than expected.
To establish these expectations, we first estimate the class-conditional relevance distribution across the class samples in the training set. We then identify outlier components whose relevance on high-confidence non-class samples deviates significantly from this distribution. These components are flagged automatically, and their associated concepts are manually inspected.
This procedure reveals over 220 failure cases of four recurrent modes: polysemous prompts (e.g., ``crane'' as bird or machine), compound nouns (e.g., ``Egyptian cat''), embedded typography in images, and spurious visual correlations.
All identified misaligned samples, where a flagged component fires strongly form the benchmark evaluation set, with examples for each failure mode provided in \cref{app:sec:robustness}. 

We quantify robustness using two metrics: the AUROC for distinguishing true class images from spurious distractors (``spurious AUC''), and from unrelated ImageNet classes (``valid AUC''). Spurious distractors include both automatically discovered cases and 50 known dataset-specific artifacts from \citet{neuhaus2023spurious}. Results are shown in \cref{fig:experiments:benchmark}.
We compare several probing strategies: (i) text-based probing using short class labels, (ii) enriched prompts via templating and detailed descriptions, and (iii) linear classifiers trained directly on visual embeddings. While CLIP exhibits greater robustness to spurious correlations compared to linear classifiers, it performs significantly worse in the presence of semantic ambiguity and typographic artifacts. Larger CLIP variants consistently show improved performance across both valid and spurious AUC.
Prompt enrichment improves robustness, particularly for typographic distractors, though gains vary across failure types. Notably, template-based prompting (e.g., ``an image of\dots'') is effective against visual typography but less so for resolving polysemy.
Detailed results and uncertainty estimates are provided in \cref{app:sec:robustness}.

\subsection{Case study: spurious correlation analysis in melanoma detection  (Q4)}
\label{sec:exp:medical}
Robustness and safety are particularly critical in medical applications. Recent works have introduced CLIP-based models for skin lesion classification, including melanoma detection from dermoscopic images~\cite{kim2024transparent,yang2024a}. While foundation models like CLIP offer rich feature representations, they may also encode complex spurious correlations that compromise reliability when used as feature extractors.

\begin{figure}[t]
    \centering
    \includegraphics[width=0.99\linewidth]{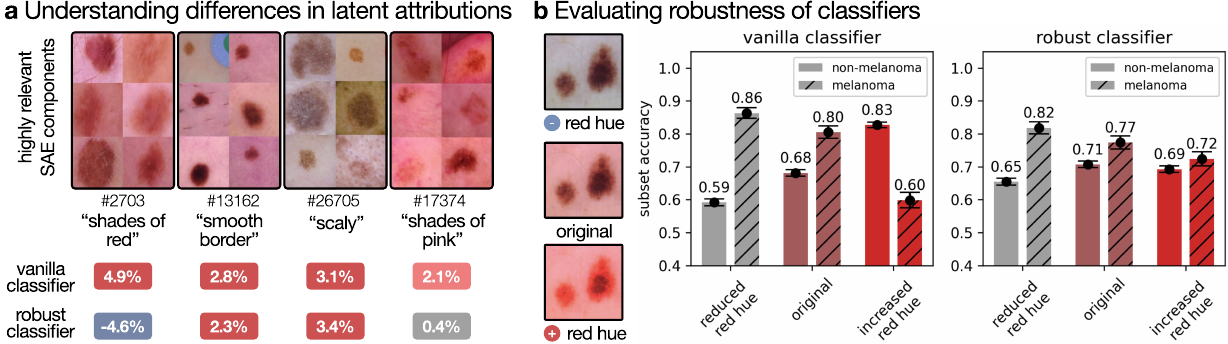}
    \caption{Melanoma detection case study.
    a) Among the most relevant SAE components for the non-melanoma class, we identify a component encoding a ``red hue’’ background artifact that strongly influences predictions when training a linear classifier on image embeddings. Training a more robust classifier on semantically augmented embeddings reduces this reliance.
    b) When systematically adding or removing red coloration from input images, the robust classifier maintains more stable accuracies on both melanoma and non-melanoma samples, demonstrating improved resilience to spurious correlations.
    Error bars correspond to the \gls{sem}.}
    \label{fig:experiments:skin_lesion}
\end{figure}

We apply our interpretability framework to WhyLesion-CLIP by training a linear classifier on image embeddings to differentiate melanoma from non-melanoma samples on the ISIC 2019 dataset. As illustrated in \cref{fig:experiments:skin_lesion}, two of the most influential SAE components encode red or pink hue background artifacts, contributing a mean relevance of 4.9\,\% and 2.8\,\% to the model’s output, respectively. When red coloration is artificially manipulated in images, classification performance on the test set shifts significantly: melanoma accuracy drops from 80.5\,\% to 59.9\,\%, while non-melanoma accuracy increases from 68.1\,\% to 82.8\,\%, confirming reliance on the red hue as a spurious cue for non-melanoma predictions. Notably, all \glspl{sem} for accuracy scores depicted in \cref{fig:experiments:skin_lesion} are below 2.4\,\%.

To mitigate this shortcut, we train a robust linear classifier on semantically augmented embeddings $\boldsymbol{x}$ by adding random signals along the latent direction associated with the red hue component $\#2703$. Specifically, we estimate the latent feature direction between strongly and weakly activating samples, following the procedure from \cite{pahde2025navigating} (see \cref{app:sec:medical_case_study} for details). This augmentation significantly improves robustness: melanoma accuracy now degrades by $(5.1\pm2.0)$\,\% under red hue increase, compared to $(20.6\pm2.1)$\,\% without.
Both color-encoding components become less relevant, with $\#2703$ now indicating melanoma instead, which is valid if corresponding specifically to the lesion~\cite{muinonen2018amelanotic}.

\section{Conclusion}
We present the first holistic framework for instance-wise attribution of sparse, interpretable components in CLIP models, enabling a dual perspective on model behavior: understanding \emph{what} concepts are encoded, and also \emph{how} they influence predictions.
By combining attributions with alignment to expected semantics, our method automatically identifies reliance on both spurious features (e.g., red backgrounds in melanoma detection) and surprising concepts (e.g. twin-like objects), potentially corresponding to novel patterns in the data. Applied across multiple CLIP variants and a real-world medical use case, we highlight persistent failure modes, including polysemous prompts, compound nouns, visual typography, and spurious correlations. 
A key advantage of including latent attributions is their simplicity: based on activation and output gradient, it is easy to implement, scale, and adopt.
Our findings emphasize the importance of mechanistic interpretability methods that move beyond global summaries to support the reliable and safe deployment of multimodal foundation models.

\textbf{Limitations:}
Throughout our experiments, we focus on class token embeddings from the final transformer block, which are most predictive, but may miss relevant dynamics in earlier layers. While we use \glspl{sae} for extracting interpretable components, alternative sparse dictionary learning methods could be applied; however, challenges like polysemanticity or hidden latent concepts may remain. Lastly, our semantic alignment relies on CLIP-Mobile-S2, which may itself be affected by spurious correlations or ambiguous language.

\textbf{Acknowledgements:}
This work was supported by
the Federal Ministry of Education and Research (BMBF) as grant BIFOLD (01IS18025A, 01IS180371I);
the European Union’s Horizon Europe research and innovation programme (EU Horizon Europe) as grants [ACHILLES (101189689), TEMA (101093003)];
and the German Research Foundation (DFG) as research unit DeSBi [KI-FOR 5363] (459422098).

\bibliography{egbib}

\begin{thebibliography}{55}
\providecommand{\natexlab}[1]{#1}
\providecommand{\url}[1]{\texttt{#1}}
\expandafter\ifx\csname urlstyle\endcsname\relax
  \providecommand{\doi}[1]{doi: #1}\else
  \providecommand{\doi}{doi: \begingroup \urlstyle{rm}\Url}\fi

\bibitem[Achtibat et~al.(2023)Achtibat, Dreyer, Eisenbraun, Bosse, Wiegand, Samek, and Lapuschkin]{achtibat2023attribution}
R.~Achtibat, M.~Dreyer, I.~Eisenbraun, S.~Bosse, T.~Wiegand, W.~Samek, and S.~Lapuschkin.
\newblock From attribution maps to human-understandable explanations through concept relevance propagation.
\newblock \emph{Nature Machine Intelligence}, 5\penalty0 (9):\penalty0 1006--1019, 2023.

\bibitem[Ahn et~al.(2024)Ahn, Kim, and Kim]{ahn2024unified}
Y.~H. Ahn, H.~B. Kim, and S.~T. Kim.
\newblock Www: a unified framework for explaining what where and why of neural networks by interpretation of neuron concepts.
\newblock In \emph{Proceedings of the IEEE/CVF Conference on Computer Vision and Pattern Recognition}, pages 10968--10977, 2024.

\bibitem[Bach et~al.(2015)Bach, Binder, Montavon, Klauschen, M{\"u}ller, and Samek]{bach2015pixel}
S.~Bach, A.~Binder, G.~Montavon, F.~Klauschen, K.-R. M{\"u}ller, and W.~Samek.
\newblock On pixel-wise explanations for non-linear classifier decisions by layer-wise relevance propagation.
\newblock \emph{PloS one}, 10\penalty0 (7):\penalty0 e0130140, 2015.

\bibitem[Belrose et~al.(2023)Belrose, Furman, Smith, Halawi, Ostrovsky, McKinney, Biderman, and Steinhardt]{belrose2023eliciting}
N.~Belrose, Z.~Furman, L.~Smith, D.~Halawi, I.~Ostrovsky, L.~McKinney, S.~Biderman, and J.~Steinhardt.
\newblock Eliciting latent predictions from transformers with the tuned lens.
\newblock \emph{arXiv preprint arXiv:2303.08112}, 2023.

\bibitem[Bhalla et~al.(2024)Bhalla, Oesterling, Srinivas, Calmon, and Lakkaraju]{bhalla2024interpreting}
U.~Bhalla, A.~Oesterling, S.~Srinivas, F.~Calmon, and H.~Lakkaraju.
\newblock Interpreting {CLIP} with sparse linear concept embeddings (spli{CE}).
\newblock In \emph{The Thirty-eighth Annual Conference on Neural Information Processing Systems}, 2024.
\newblock URL \url{https://openreview.net/forum?id=7UyBKTFrtd}.

\bibitem[Bloom and Lin(2024)]{bloom2024understanding}
J.~Bloom and J.~Lin.
\newblock Understanding sae features with the logit lens.
\newblock In \emph{AI Alignment Forum}, pages 24--25, 2024.

\bibitem[Bricken et~al.(2023)Bricken, Templeton, Batson, Chen, Jermyn, Conerly, Turner, Anil, Denison, Askell, et~al.]{bricken2023towards}
T.~Bricken, A.~Templeton, J.~Batson, B.~Chen, A.~Jermyn, T.~Conerly, N.~Turner, C.~Anil, C.~Denison, A.~Askell, et~al.
\newblock Towards monosemanticity: Decomposing language models with dictionary learning.
\newblock \emph{Transformer Circuits Thread}, 2, 2023.

\bibitem[Codella et~al.(2018)Codella, Gutman, Celebi, Helba, Marchetti, Dusza, Kalloo, Liopyris, Mishra, Kittler, et~al.]{codella2018skin}
N.~C. Codella, D.~Gutman, M.~E. Celebi, B.~Helba, M.~A. Marchetti, S.~W. Dusza, A.~Kalloo, K.~Liopyris, N.~Mishra, H.~Kittler, et~al.
\newblock Skin lesion analysis toward melanoma detection: A challenge at the 2017 international symposium on biomedical imaging (isbi), hosted by the international skin imaging collaboration (isic).
\newblock In \emph{2018 IEEE 15th international symposium on biomedical imaging (ISBI 2018)}, pages 168--172. IEEE, 2018.

\bibitem[Collaboration et~al.(2024)]{international2024slice}
I.~S.~I. Collaboration et~al.
\newblock Slice-3d 2024 challenge dataset.
\newblock \emph{International Skin Imaging Collaboration}, 10, 2024.

\bibitem[Dorszewski et~al.(2025)Dorszewski, T{\v{e}}tkov{\'a}, Jenssen, Hansen, and Wickstr{\o}m]{dorszewski2025colors}
T.~Dorszewski, L.~T{\v{e}}tkov{\'a}, R.~Jenssen, L.~K. Hansen, and K.~K. Wickstr{\o}m.
\newblock From colors to classes: Emergence of concepts in vision transformers.
\newblock \emph{arXiv preprint arXiv:2503.24071}, 2025.

\bibitem[Dreyer et~al.(2024)Dreyer, Achtibat, Samek, and Lapuschkin]{dreyer2024understanding}
M.~Dreyer, R.~Achtibat, W.~Samek, and S.~Lapuschkin.
\newblock Understanding the (extra-) ordinary: Validating deep model decisions with prototypical concept-based explanations.
\newblock In \emph{Proceedings of the IEEE/CVF Conference on Computer Vision and Pattern Recognition Workshops}, pages 3491--3501, 2024.

\bibitem[Dreyer et~al.(2025)Dreyer, Berend, Labarta, Vielhaben, Wiegand, Lapuschkin, and Samek]{dreyer2025mechanistic}
M.~Dreyer, J.~Berend, T.~Labarta, J.~Vielhaben, T.~Wiegand, S.~Lapuschkin, and W.~Samek.
\newblock Mechanistic understanding and validation of large ai models with semanticlens.
\newblock \emph{arXiv preprint arXiv:2501.05398}, 2025.

\bibitem[Elhage et~al.(2022)Elhage, Hume, Olsson, Schiefer, Henighan, Kravec, Hatfield-Dodds, Lasenby, Drain, Chen, et~al.]{elhage2022toymodelssuperposition}
N.~Elhage, T.~Hume, C.~Olsson, N.~Schiefer, T.~Henighan, S.~Kravec, Z.~Hatfield-Dodds, R.~Lasenby, D.~Drain, C.~Chen, et~al.
\newblock Toy models of superposition.
\newblock \emph{arXiv preprint arXiv:2209.10652}, 2022.

\bibitem[Fang et~al.(2024)Fang, Jose, Jain, Schmidt, Toshev, and Shankar]{fang2024data}
A.~Fang, A.~M. Jose, A.~Jain, L.~Schmidt, A.~T. Toshev, and V.~Shankar.
\newblock Data filtering networks.
\newblock In \emph{The Twelfth International Conference on Learning Representations}, 2024.
\newblock URL \url{https://openreview.net/forum?id=KAk6ngZ09F}.

\bibitem[Fel et~al.(2023{\natexlab{a}})Fel, Boutin, B{\'e}thune, Cad{\`e}ne, Moayeri, And{\'e}ol, Chalvidal, and Serre]{fel2023holistic}
T.~Fel, V.~Boutin, L.~B{\'e}thune, R.~Cad{\`e}ne, M.~Moayeri, L.~And{\'e}ol, M.~Chalvidal, and T.~Serre.
\newblock A holistic approach to unifying automatic concept extraction and concept importance estimation.
\newblock \emph{Advances in Neural Information Processing Systems}, 36:\penalty0 54805--54818, 2023{\natexlab{a}}.

\bibitem[Fel et~al.(2023{\natexlab{b}})Fel, Picard, Bethune, Boissin, Vigouroux, Colin, Cad{\`e}ne, and Serre]{fel2023craft}
T.~Fel, A.~Picard, L.~Bethune, T.~Boissin, D.~Vigouroux, J.~Colin, R.~Cad{\`e}ne, and T.~Serre.
\newblock Craft: Concept recursive activation factorization for explainability.
\newblock In \emph{Proceedings of the IEEE/CVF Conference on Computer Vision and Pattern Recognition}, pages 2711--2721, 2023{\natexlab{b}}.

\bibitem[Gadre et~al.(2023)Gadre, Ilharco, Fang, Hayase, Smyrnis, Nguyen, Marten, Wortsman, Ghosh, Zhang, et~al.]{gadre2023datacomp}
S.~Y. Gadre, G.~Ilharco, A.~Fang, J.~Hayase, G.~Smyrnis, T.~Nguyen, R.~Marten, M.~Wortsman, D.~Ghosh, J.~Zhang, et~al.
\newblock Datacomp: In search of the next generation of multimodal datasets.
\newblock \emph{Advances in Neural Information Processing Systems}, 36:\penalty0 27092--27112, 2023.

\bibitem[Gandelsman et~al.(2024)Gandelsman, Efros, and Steinhardt]{gandelsman2024interpreting}
Y.~Gandelsman, A.~A. Efros, and J.~Steinhardt.
\newblock Interpreting {CLIP}'s image representation via text-based decomposition.
\newblock In \emph{The Twelfth International Conference on Learning Representations}, 2024.
\newblock URL \url{https://openreview.net/forum?id=5Ca9sSzuDp}.

\bibitem[Gandelsman et~al.(2025)Gandelsman, Efros, and Steinhardt]{gandelsman2025interpreting}
Y.~Gandelsman, A.~A. Efros, and J.~Steinhardt.
\newblock Interpreting the second-order effects of neurons in {CLIP}.
\newblock In \emph{The Thirteenth International Conference on Learning Representations}, 2025.
\newblock URL \url{https://openreview.net/forum?id=GPDcvoFGOL}.

\bibitem[Gao et~al.(2025)Gao, la~Tour, Tillman, Goh, Troll, Radford, Sutskever, Leike, and Wu]{gao2025scaling}
L.~Gao, T.~D. la~Tour, H.~Tillman, G.~Goh, R.~Troll, A.~Radford, I.~Sutskever, J.~Leike, and J.~Wu.
\newblock Scaling and evaluating sparse autoencoders.
\newblock In \emph{The Thirteenth International Conference on Learning Representations}, 2025.
\newblock URL \url{https://openreview.net/forum?id=tcsZt9ZNKD}.

\bibitem[Hern{\'a}ndez-P{\'e}rez et~al.(2024)Hern{\'a}ndez-P{\'e}rez, Combalia, Podlipnik, Codella, Rotemberg, Halpern, Reiter, Carrera, Barreiro, Helba, et~al.]{hernandez2024bcn20000}
C.~Hern{\'a}ndez-P{\'e}rez, M.~Combalia, S.~Podlipnik, N.~C. Codella, V.~Rotemberg, A.~C. Halpern, O.~Reiter, C.~Carrera, A.~Barreiro, B.~Helba, et~al.
\newblock Bcn20000: Dermoscopic lesions in the wild.
\newblock \emph{Scientific data}, 11\penalty0 (1):\penalty0 641, 2024.

\bibitem[Huben et~al.(2024)Huben, Cunningham, Smith, Ewart, and Sharkey]{huben2024sparse}
R.~Huben, H.~Cunningham, L.~R. Smith, A.~Ewart, and L.~Sharkey.
\newblock Sparse autoencoders find highly interpretable features in language models.
\newblock In \emph{The Twelfth International Conference on Learning Representations}, 2024.
\newblock URL \url{https://openreview.net/forum?id=F76bwRSLeK}.

\bibitem[Iglewicz and Hoaglin(1993)]{iglewicz1993volume}
B.~Iglewicz and D.~C. Hoaglin.
\newblock \emph{Volume 16: how to detect and handle outliers}.
\newblock Quality Press, 1993.

\bibitem[Ilharco et~al.(2021)Ilharco, Wortsman, Wightman, Gordon, Carlini, Taori, Dave, Shankar, Namkoong, Miller, Hajishirzi, Farhadi, and Schmidt]{ilharco_gabriel_2021_5143773}
G.~Ilharco, M.~Wortsman, R.~Wightman, C.~Gordon, N.~Carlini, R.~Taori, A.~Dave, V.~Shankar, H.~Namkoong, J.~Miller, H.~Hajishirzi, A.~Farhadi, and L.~Schmidt.
\newblock Openclip, July 2021.
\newblock URL \url{https://doi.org/10.5281/zenodo.5143773}.

\bibitem[Joseph et~al.(2025)Joseph, Suresh, Goldfarb, Hufe, Gandelsman, Graham, Bzdok, Samek, and Richards]{joseph2025steering}
S.~Joseph, P.~Suresh, E.~Goldfarb, L.~Hufe, Y.~Gandelsman, R.~Graham, D.~Bzdok, W.~Samek, and B.~A. Richards.
\newblock Steering clip's vision transformer with sparse autoencoders.
\newblock In \emph{Mechanistic Interpretability for Vision at CVPR 2025 (Non-proceedings Track)}, 2025.

\bibitem[Kim et~al.(2024)Kim, Gadgil, DeGrave, Omiye, Cai, Daneshjou, and Lee]{kim2024transparent}
C.~Kim, S.~U. Gadgil, A.~J. DeGrave, J.~A. Omiye, Z.~R. Cai, R.~Daneshjou, and S.-I. Lee.
\newblock Transparent medical image ai via an image–text foundation model grounded in medical literature.
\newblock \emph{Nature Medicine}, 2024.
\newblock \doi{10.1038/s41591-024-02887-x}.
\newblock URL \url{https://doi.org/10.1038/s41591-024-02887-x}.

\bibitem[Krizhevsky et~al.(2012)Krizhevsky, Sutskever, and Hinton]{krizhevsky2012imagenet}
A.~Krizhevsky, I.~Sutskever, and G.~E. Hinton.
\newblock Imagenet classification with deep convolutional neural networks.
\newblock \emph{Advances in neural information processing systems}, 25, 2012.

\bibitem[Kumar et~al.(2024)Kumar, Ghosh, Sakshi, Tyagi, and Manocha]{kumar2024vision}
S.~Kumar, S.~Ghosh, S.~Sakshi, U.~Tyagi, and D.~Manocha.
\newblock Do vision-language models understand compound nouns?
\newblock In \emph{Proceedings of the 2024 Conference of the North American Chapter of the Association for Computational Linguistics: Human Language Technologies (Volume 2: Short Papers)}, pages 519--527, 2024.

\bibitem[Li et~al.(2023)Li, Li, Savarese, and Hoi]{li2023blip}
J.~Li, D.~Li, S.~Savarese, and S.~Hoi.
\newblock Blip-2: Bootstrapping language-image pre-training with frozen image encoders and large language models.
\newblock In \emph{International conference on machine learning}, pages 19730--19742. PMLR, 2023.

\bibitem[Lim et~al.(2025)Lim, Choi, Choo, and Schneider]{lim2025sparse}
H.~Lim, J.~Choi, J.~Choo, and S.~Schneider.
\newblock Sparse autoencoders reveal selective remapping of visual concepts during adaptation.
\newblock In \emph{The Thirteenth International Conference on Learning Representations}, 2025.
\newblock URL \url{https://openreview.net/forum?id=imT03YXlG2}.

\bibitem[Loshchilov and Hutter(2019)]{loshchilovdecoupled}
I.~Loshchilov and F.~Hutter.
\newblock Decoupled weight decay regularization.
\newblock In \emph{International Conference on Learning Representations}, 2019.

\bibitem[Marks et~al.(2025)Marks, Rager, Michaud, Belinkov, Bau, and Mueller]{marks2025sparse}
S.~Marks, C.~Rager, E.~J. Michaud, Y.~Belinkov, D.~Bau, and A.~Mueller.
\newblock Sparse feature circuits: Discovering and editing interpretable causal graphs in language models.
\newblock In \emph{The Thirteenth International Conference on Learning Representations}, 2025.
\newblock URL \url{https://openreview.net/forum?id=I4e82CIDxv}.

\bibitem[Muinonen-Martin et~al.(2018)Muinonen-Martin, O'Shea, and Newton-Bishop]{muinonen2018amelanotic}
A.~J. Muinonen-Martin, S.~J. O'Shea, and J.~Newton-Bishop.
\newblock Amelanotic melanoma.
\newblock \emph{BMJ (Clinical research ed.)}, 360:\penalty0 k826, 2018.

\bibitem[Nanda(2023)]{nanda2023attribution}
N.~Nanda.
\newblock Attribution patching: Activation patching at industrial scale.
\newblock \emph{URL: https://www. neelnanda. io/mechanistic-interpretability/attribution-patching}, 2023.

\bibitem[Neuhaus et~al.(2023)Neuhaus, Augustin, Boreiko, and Hein]{neuhaus2023spurious}
Y.~Neuhaus, M.~Augustin, V.~Boreiko, and M.~Hein.
\newblock Spurious features everywhere-large-scale detection of harmful spurious features in imagenet.
\newblock In \emph{Proceedings of the IEEE/CVF International Conference on Computer Vision}, pages 20235--20246, 2023.

\bibitem[Olshausen and Field(1997)]{olshausen1997sparse}
B.~A. Olshausen and D.~J. Field.
\newblock Sparse coding with an overcomplete basis set: A strategy employed by v1?
\newblock \emph{Vision research}, 37\penalty0 (23):\penalty0 3311--3325, 1997.

\bibitem[Pach et~al.(2025)Pach, Karthik, Bouniot, Belongie, and Akata]{pach2025sparse}
M.~Pach, S.~Karthik, Q.~Bouniot, S.~Belongie, and Z.~Akata.
\newblock Sparse autoencoders learn monosemantic features in vision-language models.
\newblock \emph{arXiv preprint arXiv:2504.02821}, 2025.

\bibitem[Pahde et~al.(2025)Pahde, Dreyer, Weckbecker, Weber, Anders, Wiegand, Samek, and Lapuschkin]{pahde2025navigating}
F.~Pahde, M.~Dreyer, M.~Weckbecker, L.~Weber, C.~J. Anders, T.~Wiegand, W.~Samek, and S.~Lapuschkin.
\newblock Navigating neural space: Revisiting concept activation vectors to overcome directional divergence.
\newblock In \emph{The Thirteenth International Conference on Learning Representations}, 2025.
\newblock URL \url{https://openreview.net/forum?id=Q95MaWfF4e}.

\bibitem[Radford et~al.(2021)Radford, Kim, Hallacy, Ramesh, Goh, Agarwal, Sastry, Askell, Mishkin, Clark, et~al.]{radford2021learning}
A.~Radford, J.~W. Kim, C.~Hallacy, A.~Ramesh, G.~Goh, S.~Agarwal, G.~Sastry, A.~Askell, P.~Mishkin, J.~Clark, et~al.
\newblock Learning transferable visual models from natural language supervision.
\newblock In \emph{International conference on machine learning}, pages 8748--8763. PMLR, 2021.

\bibitem[Rao et~al.(2024)Rao, Mahajan, B{\"o}hle, and Schiele]{rao2024discover}
S.~Rao, S.~Mahajan, M.~B{\"o}hle, and B.~Schiele.
\newblock Discover-then-name: Task-agnostic concept bottlenecks via automated concept discovery.
\newblock In \emph{European Conference on Computer Vision}, pages 444--461. Springer, 2024.

\bibitem[Ridnik et~al.(2021)Ridnik, Ben-Baruch, Noy, and Zelnik-Manor]{ridnik2021imagenetk}
T.~Ridnik, E.~Ben-Baruch, A.~Noy, and L.~Zelnik-Manor.
\newblock Imagenet-21k pretraining for the masses.
\newblock In \emph{Thirty-fifth Conference on Neural Information Processing Systems Datasets and Benchmarks Track (Round 1)}, 2021.
\newblock URL \url{https://openreview.net/forum?id=Zkj_VcZ6ol}.

\bibitem[Rotemberg et~al.(2021)Rotemberg, Kurtansky, Betz-Stablein, Caffery, Chousakos, Codella, Combalia, Dusza, Guitera, Gutman, et~al.]{rotemberg2021patient}
V.~Rotemberg, N.~Kurtansky, B.~Betz-Stablein, L.~Caffery, E.~Chousakos, N.~Codella, M.~Combalia, S.~Dusza, P.~Guitera, D.~Gutman, et~al.
\newblock A patient-centric dataset of images and metadata for identifying melanomas using clinical context.
\newblock \emph{Scientific data}, 8\penalty0 (1):\penalty0 34, 2021.

\bibitem[Schuhmann et~al.(2021)Schuhmann, Vencu, Beaumont, Kaczmarczyk, Mullis, Katta, Coombes, Jitsev, and Komatsuzaki]{schuhmann2021laion}
C.~Schuhmann, R.~Vencu, R.~Beaumont, R.~Kaczmarczyk, C.~Mullis, A.~Katta, T.~Coombes, J.~Jitsev, and A.~Komatsuzaki.
\newblock Laion-400m: Open dataset of clip-filtered 400 million image-text pairs.
\newblock \emph{arXiv preprint arXiv:2111.02114}, 2021.

\bibitem[Schuhmann et~al.(2022)Schuhmann, Beaumont, Vencu, Gordon, Wightman, Cherti, Coombes, Katta, Mullis, Wortsman, et~al.]{schuhmann2022laion}
C.~Schuhmann, R.~Beaumont, R.~Vencu, C.~Gordon, R.~Wightman, M.~Cherti, T.~Coombes, A.~Katta, C.~Mullis, M.~Wortsman, et~al.
\newblock Laion-5b: An open large-scale dataset for training next generation image-text models.
\newblock \emph{Advances in neural information processing systems}, 35:\penalty0 25278--25294, 2022.

\bibitem[Selvaraju et~al.(2017)Selvaraju, Cogswell, Das, Vedantam, Parikh, and Batra]{selvaraju2017grad}
R.~R. Selvaraju, M.~Cogswell, A.~Das, R.~Vedantam, D.~Parikh, and D.~Batra.
\newblock Grad-cam: Visual explanations from deep networks via gradient-based localization.
\newblock In \emph{Proceedings of the IEEE international conference on computer vision}, pages 618--626, 2017.

\bibitem[Shrikumar et~al.(2017)Shrikumar, Greenside, and Kundaje]{shrikumar2017learning}
A.~Shrikumar, P.~Greenside, and A.~Kundaje.
\newblock Learning important features through propagating activation differences.
\newblock In \emph{International conference on machine learning}, pages 3145--3153. PMlR, 2017.

\bibitem[Syed et~al.(2024)Syed, Rager, and Conmy]{syed2024attribution}
A.~Syed, C.~Rager, and A.~Conmy.
\newblock Attribution patching outperforms automated circuit discovery.
\newblock In \emph{Proceedings of the 7th BlackboxNLP Workshop: Analyzing and Interpreting Neural Networks for NLP}, pages 407--416, 2024.

\bibitem[Thasarathan et~al.(2025)Thasarathan, Forsyth, Fel, Kowal, and Derpanis]{thasarathan2025universal}
H.~Thasarathan, J.~Forsyth, T.~Fel, M.~Kowal, and K.~Derpanis.
\newblock Universal sparse autoencoders: Interpretable cross-model concept alignment.
\newblock \emph{arXiv preprint arXiv:2502.03714}, 2025.

\bibitem[Tschandl et~al.(2018)Tschandl, Rosendahl, and Kittler]{tschandl2018ham10000}
P.~Tschandl, C.~Rosendahl, and H.~Kittler.
\newblock The ham10000 dataset, a large collection of multi-source dermatoscopic images of common pigmented skin lesions.
\newblock \emph{Scientific data}, 5\penalty0 (1):\penalty0 1--9, 2018.

\bibitem[Vasu et~al.(2024)Vasu, Pouransari, Faghri, Vemulapalli, and Tuzel]{vasu2024mobileclip}
P.~K.~A. Vasu, H.~Pouransari, F.~Faghri, R.~Vemulapalli, and O.~Tuzel.
\newblock Mobileclip: Fast image-text models through multi-modal reinforced training.
\newblock In \emph{Proceedings of the IEEE/CVF Conference on Computer Vision and Pattern Recognition}, pages 15963--15974, 2024.

\bibitem[Vig et~al.(2020)Vig, Gehrmann, Belinkov, Qian, Nevo, Singer, and Shieber]{vig2020investigating}
J.~Vig, S.~Gehrmann, Y.~Belinkov, S.~Qian, D.~Nevo, Y.~Singer, and S.~Shieber.
\newblock Investigating gender bias in language models using causal mediation analysis.
\newblock \emph{Advances in neural information processing systems}, 33:\penalty0 12388--12401, 2020.

\bibitem[Wang et~al.(2024)Wang, Lin, Chen, Schmidt, Han, and Zhang]{wang2024a}
Q.~Wang, Y.~Lin, Y.~Chen, L.~Schmidt, B.~Han, and T.~Zhang.
\newblock A sober look at the robustness of {CLIP}s to spurious features.
\newblock In \emph{The Thirty-eighth Annual Conference on Neural Information Processing Systems}, 2024.
\newblock URL \url{https://openreview.net/forum?id=wWyumwEYV8}.

\bibitem[Yang et~al.(2024)Yang, Gandhi, Wang, Wu, Yao, Callison-Burch, Gee, and Yatskar]{yang2024a}
Y.~Yang, M.~Gandhi, Y.~Wang, Y.~Wu, M.~S. Yao, C.~Callison-Burch, J.~Gee, and M.~Yatskar.
\newblock A textbook remedy for domain shifts: Knowledge priors for medical image analysis.
\newblock In \emph{The Thirty-eighth Annual Conference on Neural Information Processing Systems}, 2024.
\newblock URL \url{https://openreview.net/forum?id=STrpbhrvt3}.

\bibitem[Yun et~al.(2023)Yun, Bhalla, Pavlick, and Sun]{yun2023do}
T.~Yun, U.~Bhalla, E.~Pavlick, and C.~Sun.
\newblock Do vision-language pretrained models learn composable primitive concepts?
\newblock \emph{Transactions on Machine Learning Research}, 2023.
\newblock ISSN 2835-8856.
\newblock URL \url{https://openreview.net/forum?id=YwNrPLjHSL}.

\bibitem[Zhu et~al.(2024)Zhu, Chen, Shen, Li, and Elhoseiny]{zhu2024minigpt}
D.~Zhu, J.~Chen, X.~Shen, X.~Li, and M.~Elhoseiny.
\newblock Mini{GPT}-4: Enhancing vision-language understanding with advanced large language models.
\newblock In \emph{The Twelfth International Conference on Learning Representations}, 2024.
\newblock URL \url{https://openreview.net/forum?id=1tZbq88f27}.

\end{thebibliography}

\newpage
\appendix

\section*{Appendix}

The appendix provides more methodological and experimental details for the main manuscript, begging with details on the experimental settings in~\cref{app:sec:experimental_settings}, followed by additional descriptions and results on attribution faithfulness evaluations in~\cref{app:sec:latent_attributions}, interpretability and diversity analysis of \gls{sae} components in~\cref{app:sec:sae_interpretability}, CLIP robustness evaluations in~\cref{app:sec:robustness}, and the medical case study in~\cref{app:sec:medical_case_study}.

\section{Experimental settings}
\label{app:sec:experimental_settings}
This section provides more details on models, datasets, SAE training configurations, and computational resources for our experiments.

\paragraph{Models}  
We evaluate several CLIP variants: ViT-B/32 trained on LAION-400M~\cite{schuhmann2021laion}, LAION-2B~\cite{schuhmann2022laion}, and DataComp-(M, XL)~\cite{gadre2023datacomp}; ViT-B/16 and ViT-L/14 on DataComp-XL; and ViT-H/14 trained on DFN-5B~\cite{fang2024data}. For semantic alignment scores (\cref{eq:background:labeling}), we use Mobile-CLIP-S2~\cite{vasu2024mobileclip}. For the medical setting, we evaluate on WhyLesion-CLIP~\cite{yang2024a}.

\paragraph{Datasets}
We use ImageNet-1k~\cite{krizhevsky2012imagenet} for experiments in \cref{sec:exp:faithfulness,sec:exp:sae,sec:exp:robustness}.
Semantic alignment scores (\cref{eq:background:labeling}) are evaluated against the ImageNet-21k class names~\cite{ridnik2021imagenetk}.
For the medical analysis in \cref{sec:exp:medical}, we use ISIC datasets of 2019~\cite{tschandl2018ham10000,codella2018skin,hernandez2024bcn20000} for the main analysis, and also 2020~\cite{rotemberg2021patient} and 2024~\cite{international2024slice} for \gls{sae} training.

\paragraph{Sparse autoencoder}
We train top-$k$ \glspl{sae}~\cite{gao2025scaling} with $k=64$ and 30{,}000 components. For general vision CLIP models, training is performed on the ImageNet-1k training set. For the medical task, we use a combined dataset of ISIC 2019, 2020, and 2024.  
For the \glspl{sae} as described by \cref{eq:background:sae}, we enforce latent vectors $\mathbf{v}_j$ to have a norm of one:
\begin{equation}
    \|\mathbf{v}_j\| = 1\,.
\end{equation}

For ImageNet \glspl{sae}, we train with AdamW~\cite{loshchilovdecoupled} and a learning rate of $1\cdot10^{-6}$ for 30 epochs, reducing it by a factor of 10 after epochs 24 and 28. Each epoch is performed over 10\,\% of the full train dataset (randomly sampled). For the medical setting, we train for 25 epochs with a learning rate of $5\cdot10^{-5}$, decayed at epochs 17 and 23.

The loss for training corresponds to an MSE loss between original $\boldsymbol{x}_j$ and reconstructed normalized embeddings $\hat{\boldsymbol{x}}_j$ for token $i$:
\begin{equation}
    L_\text{MSE} = L_\text{MSE}^\text{cls} + \frac{1}{m}\sum_j^m L_\text{MSE}^j\,, \quad \text{with} \quad L_\text{MSE}^i = \left\|\frac{{\boldsymbol{x}}_j}{\|{\boldsymbol{x}}_j\|} - \frac{\hat{\boldsymbol{x}}_j}{\|\hat{\boldsymbol{x}}_j\|}\right\|^2 \,,
\end{equation}
where $m$ is the number of spatial tokens.
Note, that in the end, the \gls{sae} is used to reconstruct the class token embedding $\boldsymbol{x}_\text{cls}$ only, but can, in principle, also reconstruct the spatial tokens.

\paragraph{Computational resources}
\label{app:sec:experimental_settings:computational_ressources}

Training of \glspl{sae} has been performed on a GPU cluster with either one of four A100 40GB or Tesla V100 32GB GPUs by NVIDIA per node.
Training of the ViT-H/14 SAE takes about 12 hours (A100), ViT-L/14 16 hours (Tesla V100), ViT-B/16 6 hours (Tesla V100), ViT-B/32 3 hours (Tesla V100).
For all other experiments,
a NVIDIA Titan RTX 24GB on a local workstation was used.
Other compute intensive tasks correspond to the collection of visual embeddings and \gls{sae} component activations on the ImageNet-1k train set.
This takes about 11 hours for the ViT-H/14 and 5 hours for the ViT-L/14 model and 1.5 hours for the ViT-B/16 model.
All other experiments require less than 1 hour of compute time per model.

\section{Latent component attributions}
\label{app:sec:latent_attributions}

This section provides additional methodological details for \cref{sec:methods:attributions} and results on evaluating the faithfulness of instance-wise attributions using \gls{sae} components for \cref{sec:exp:faithfulness}.

\subsection{Approximation of activation times gradient}

We aim to compute attribution scores $R_j(\boldsymbol{x}, \mathbf{t})$ for component $j$, with output embedding $\boldsymbol{x}$ and textual prompt $\mathbf{t}$, explaining an output $y(\boldsymbol{x}, \mathbf{t})$. The relevance score is defined as
\begin{align}
\label{eq:methods:relevance}
    R_j(\boldsymbol{x}, \mathbf{t})
    = a_j\frac{\partial y(\boldsymbol{x}, \mathbf{t})}{\partial a_j}~.
\end{align}
We also know that $y(\boldsymbol{x}, \mathbf{t})$ is given by the cosine similarity between $\boldsymbol{x}$ and $\mathbf{t}$:
\begin{equation}
    \frac{\partial y(\boldsymbol{x}, \mathbf{t})}{\partial a_j} =  \frac{\partial}{\partial a_j} \frac{\boldsymbol{x}}{\|\boldsymbol{x\|}} \cdot \frac{\mathbf{t}}{\|\mathbf{t}\|}~,
\end{equation}
where $\mathbf{t}$ is independt of $a_j$. Thus, using the quotient rule, we have to solve for
\begin{equation}
\label{app:eq:solve_x_over_norm_of_x}
    \frac{\partial}{\partial a_j} \frac{\boldsymbol{x}}{\|\boldsymbol{x\|}} = \frac{\|\boldsymbol{x}\| \frac{\partial}{\partial a_j}\boldsymbol{x} - \boldsymbol{x}\frac{\partial}{\partial a_j}\|\boldsymbol{x}\|}{\|\boldsymbol{x}\|^2}~.
\end{equation}

We begin with solving for $\frac{\partial}{\partial a_j}\boldsymbol{x}$ given as
\begin{equation}
    \frac{\partial}{\partial a_j}\boldsymbol{x} = \frac{\partial}{\partial a_j} \text{LayerNorm}_{\boldsymbol{\gamma},\boldsymbol{\beta}}[\mathbf{x}]W_\text{proj}
\end{equation}

where the LayerNorm operation leads to

\begin{equation}
    \text{LayerNorm}_{\boldsymbol{\gamma},\boldsymbol{\beta}}[\mathbf{x}] = \frac{\mathbf{x} - \mu_\mathbf{x}}{\|\mathbf{x} - \mu_\mathbf{x}\|}\circ \boldsymbol{\gamma} + \boldsymbol{\beta}
\end{equation}
with average $\mu_\mathbf{x}$ over elements of $\mathbf{x}$, element-wise scaling via $\boldsymbol{\gamma}$ and bias term $\boldsymbol{\beta}$.

Applying LayerNorm to a single latent component vector $\mathbf{v}_j$ results in
\begin{equation}
    \text{LayerNorm}_{\boldsymbol{\gamma},\boldsymbol{\beta}}[\mathbf{v}_j] = \frac{\mathbf{v}_j - \mu_{\mathbf{v}_j}}{\|\mathbf{v}_j - \mu_{\mathbf{v}_j}\|}\circ \boldsymbol{\gamma} + \boldsymbol{\beta}~.
\end{equation}

Notably,
$\mathbf{x} = \sum_i^{d_\text{SAE}}a_i\mathbf{v}_i + \mathbf{b} + \boldsymbol{\varepsilon}$ 
with feature vectors $\mathbf{v}_i$ for latent SAE components, a bias and error term (see \cref{eq:background:sae}).
We now include the bias and error term in the sum, introducing $a_{d_\text{SAE}+1} = a_{d_\text{SAE}+2} = 1$ and $\mathbf{v}_{d_\text{SAE}+1} = \mathbf{b}$ and $\mathbf{v}_{d_\text{SAE}+2} = \boldsymbol{\varepsilon}$ , resulting in
\begin{equation}
    \mathbf{x} = \sum_i^{d_\text{SAE}+2}a_i\mathbf{v}_i~.
\end{equation}
We further note that
\begin{equation}
\label{app:eq:x_minus_mu}
    \mathbf{x} - \mu_\mathbf{x} = \sum_i^{d_\text{SAE}+2}a_i(\mathbf{v}_i - \mu_{\mathbf{v}_i})~.
\end{equation}

We in the following assume, that $\boldsymbol{\beta}$ is very small and neglectable, i.e., $\boldsymbol{\beta}\overset{!}{=}0$.
Then, using \cref{app:eq:x_minus_mu}, we can write
\begin{equation}
    \text{LayerNorm}_{\boldsymbol{\gamma},\boldsymbol{\beta}}[\mathbf{x}] = \sum_i a_i\text{LayerNorm}_{\boldsymbol{\gamma},\boldsymbol{\beta}}[\mathbf{v}_i] \frac{\|\mathbf{v}_i - \mu_{\mathbf{v}_i}\|}{\|\mathbf{x} - \mu_\mathbf{x}\|}
\end{equation}

By defining $\boldsymbol{v}_i = \text{LayerNorm}_{\boldsymbol{\gamma},\boldsymbol{\beta}}[\mathbf{v}_i]W_\text{proj}$, we can write
\begin{equation}
    \boldsymbol{x} = \text{LayerNorm}_{\boldsymbol{\gamma},\boldsymbol{\beta}}[\mathbf{x}]W_\text{proj}
    =  \frac{\sum_i a_i \|\mathbf{v}_i - \mu_{\mathbf{v}_i}\|\boldsymbol{v}_i}{\|\mathbf{x} - \mu_\mathbf{x}\|}~.
\end{equation}

Now, by inserting this last formulation into $\frac{\partial}{\partial a_j}\boldsymbol{x}$, we receive
\begin{equation}
    \frac{\partial}{\partial a_j}\boldsymbol{x} = \
    \frac{\partial}{\partial a_j}\frac{\sum_i a_i \|\mathbf{v}_i - \mu_{\mathbf{v}_i}\|\boldsymbol{v}_i}{\|\mathbf{x} - \mu_\mathbf{x}\|} = \frac{\|\mathbf{x} - \mu_\mathbf{x}\|\|\mathbf{v}_j - \mu_{\mathbf{v}_j}\|\boldsymbol{v}_j - \sum_i a_i \|\mathbf{v}_i - \mu_{\mathbf{v}_i}\|\boldsymbol{v}_i \frac{\partial}{\partial a_j}\|\mathbf{x} - \mu_\mathbf{x}\|}{\|\mathbf{x} - \mu_\mathbf{x}\|^2}~,
\end{equation}
where we have used the quotient rule again.
We now aim to solve for the term of $ \frac{\partial}{\partial a_j} \|\mathbf{x} - \mu_\mathbf{x}\|$ given as
\begin{equation}
    \frac{\partial}{\partial a_j} \|\mathbf{x} - \mu_\mathbf{x}\| = \frac{(\mathbf{x} - \mu_\mathbf{x})\cdot (\mathbf{v}_j - \mu_{\mathbf{v}_j})}{\|\mathbf{x} - \mu_\mathbf{x}\|}~,
\end{equation}
which,
inserted into previous result for $ \frac{\partial}{\partial a_j}\boldsymbol{x} $ gives
\begin{align}
    \frac{\partial}{\partial a_j}\boldsymbol{x} 
     &= 
     \frac{\|\mathbf{x} - \mu_\mathbf{x}\|\|\mathbf{v}_j - \mu_{\mathbf{v}_j}\|\boldsymbol{v}_j - \boldsymbol{x} (\mathbf{x} - \mu_\mathbf{x})\cdot (\mathbf{v}_j - \mu_{\mathbf{v}_j})}{\|\mathbf{x} - \mu_\mathbf{x}\|^2} \\
     &=\frac{\|\mathbf{v}_j - \mu_{\mathbf{v}_j}\|}{\|\mathbf{x} - \mu_\mathbf{x}\|}\left[\boldsymbol{v}_j - \frac{\mathbf{v}_j - \mu_{\mathbf{v}_j}}{\|\mathbf{v}_j - \mu_{\mathbf{v}_j}\|}\cdot\frac{\mathbf{x} - \mu_{\mathbf{x}}}{\|\mathbf{x} - \mu_{\mathbf{x}}\|} \boldsymbol{x} \right]~.
\end{align}

Next,
we have to solve $\frac{\partial}{\partial a_j}\|\boldsymbol{x}\|$ as in \cref{app:eq:solve_x_over_norm_of_x}:

\begin{align}
    \frac{\partial}{\partial a_j}\|\boldsymbol{x}\| &= \frac{1}{\|\boldsymbol{x}\|} \boldsymbol{x}\cdot \frac{\partial}{\partial a_j}\boldsymbol{x} \\
    &= \frac{\|\mathbf{v}_j - \mu_{\mathbf{v}_j}\|}{\|\mathbf{x} - \mu_\mathbf{x}\|}
    \left[\frac{\boldsymbol{x}}{\|\boldsymbol{x}\|} \cdot \boldsymbol{v}_j - \frac{\mathbf{v}_j - \mu_{\mathbf{v}_j}}{\|\mathbf{v}_j - \mu_{\mathbf{v}_j}\|}\cdot\frac{\mathbf{x} - \mu_{\mathbf{x}}}{\|\mathbf{x} - \mu_{\mathbf{x}}\|} \right]~.
\end{align}

Inserting previous results into \cref{app:eq:solve_x_over_norm_of_x} leads to:

\begin{align}
    \frac{\partial}{\partial a_j} \frac{\boldsymbol{x}}{\|\boldsymbol{x}\|} &= 
    \frac{\|\mathbf{v}_j - \mu_{\mathbf{v}_j}\|}{\|\mathbf{x} - \mu_\mathbf{x}\|} \frac{1}{\|\boldsymbol{x}\|} 
    \left[ 
    \boldsymbol{v}_j
    -
     \frac{\boldsymbol{x} \cdot \boldsymbol{v}_j}{\|\boldsymbol{x}\|}  \frac{\boldsymbol{x}}{\|\boldsymbol{x}\|}
    \right] \\
    &= \frac{\|\mathbf{v}_j - \mu_{\mathbf{v}_j}\|}{\|\mathbf{x} - \mu_\mathbf{x}\|} \frac{\|\boldsymbol{v}_j\|}{\|\boldsymbol{x}\|} 
    \left[ 
    \frac{\boldsymbol{v}_j}{\|\boldsymbol{v}_j\|}
    -
     \frac{\boldsymbol{x} \cdot \boldsymbol{v}_j}{\|\boldsymbol{x}\|\|\boldsymbol{v}_j\|}  \frac{\boldsymbol{x}}{\|\boldsymbol{x}\|}
    \right]~,
\end{align}
where two terms cancel out.

Finally, we can solve for the term of $R_j(\boldsymbol{x}, \mathbf{t})$:
\begin{align}
\label{eq:methods:relevance}
    R_j(\boldsymbol{x}, \mathbf{t})
    &= a_j\frac{\partial y(\boldsymbol{x}, \mathbf{t})}{\partial a_j} = a_j \frac{\partial}{\partial a_j} \frac{\boldsymbol{x}}{\|\boldsymbol{x\|}} \cdot \frac{\mathbf{t}}{\|\mathbf{t}\|}\\
    &= 
    \frac{a_j\|\mathbf{v}_j - \mu_{\mathbf{v}_j}\|}{\|\mathbf{x} - \mu_\mathbf{x}\|} \frac{\|\boldsymbol{v}_j\|}{\|\boldsymbol{x}\|} 
    \left[ 
    \frac{\boldsymbol{v}_j \cdot \mathbf{t}}{\|\boldsymbol{v}_j\|\|\mathbf{t}\|}
    -
     \frac{\boldsymbol{x} \cdot \boldsymbol{v}_j}{\|\boldsymbol{x}\|\|\boldsymbol{v}_j\|}  \frac{\boldsymbol{x}\cdot \mathbf{t}}{\|\boldsymbol{x}\|\|\mathbf{t}\|}
    \right] \\
     &= 
     a_j\frac{\|\mathbf{v}_j - \mu_{\mathbf{v}_j}\|}{\|\mathbf{x} - \mu_\mathbf{x}\|} \frac{\|\boldsymbol{v}_j\|}{\|\boldsymbol{x}\|} 
    \left[ 
    \texttt{LogitLens}_j(\mathbf{t})
    -
     \frac{\boldsymbol{x} \cdot \boldsymbol{v}_j}{\|\boldsymbol{x}\|\|\boldsymbol{v}_j\|}  y(\boldsymbol{x}, \mathbf{t})
    \right]~,
\end{align}
where we included $\texttt{LogitLens}_j(\mathbf{t})$ corresponding to the Logit Lens alignment score for component $j$ as given by \cref{eq:preliminaries:logitlens}.

\subsection{Evaluating faithfulness of attributions}
This section provides more details and results for experiments of \cref{sec:exp:faithfulness}.
We perform all evaluations on the test set of ImageNet-1k, which contains 50,000 samples with 50 images per class.

\textbf{Deletion (local):} We evaluate the first 100 classes of ImageNet-1k, using 50 samples per class. For each image, we compute attribution scores indicating the relevance of each \gls{sae} component to the cosine similarity between the visual and text embeddings of the corresponding class label. The top 10 most relevant components are sequentially set to zero activation, and we measure the reduction in output similarity at each step. The area under this deletion curve (AUC) quantifies the contribution of the identified components.

\textbf{Deletion (global):} We compute global relevance scores based on the 50 samples per class. These scores are then used to select components for deletion across all 50 samples per class, allowing us to assess how globally important components affect prediction similarity when ablated.

\textbf{Deletion (random):} We compute global relevance scores based on 500 randomly sampled images. These scores are then used to select components for deletion across all 50 samples per class, allowing us to assess how important the choice of reference samples is for a method.

\textbf{Insertion (local):} This procedure mirrors local deletion but begins with all component activations set to zero. The most relevant components are then sequentially reinserted by restoring their original activation values, enabling us to assess their cumulative contribution to the model's output.

\textbf{Baselines:}
We compare our method based on \cref{eq:methods:relevance} (Act.$\times$Grad) against Logit Lens, activation times Logit Lens (which mimicks \cref{eq:methods:relevance} without the local correction term), Energy~\cite{thasarathan2025universal} (corresponding to component activation, as the norm of latent vectors is one for our \glspl{sae}), Integrated Gradients as proposed in \cite{marks2025sparse} with 10 steps, and a random choice of activating latent componens.

\textbf{Uncertainty Estimation:}  
To estimate the uncertainty of the AUC scores, we divide the evaluation data into 9 equally sized subsets. For each subset, we compute the AUC independently. The standard error of the mean (\gls{sem}) across these AUC values is then used as the uncertainty estimate.

\textbf{Results:}  
\begin{figure}
    \centering
    \includegraphics[width=1\linewidth]{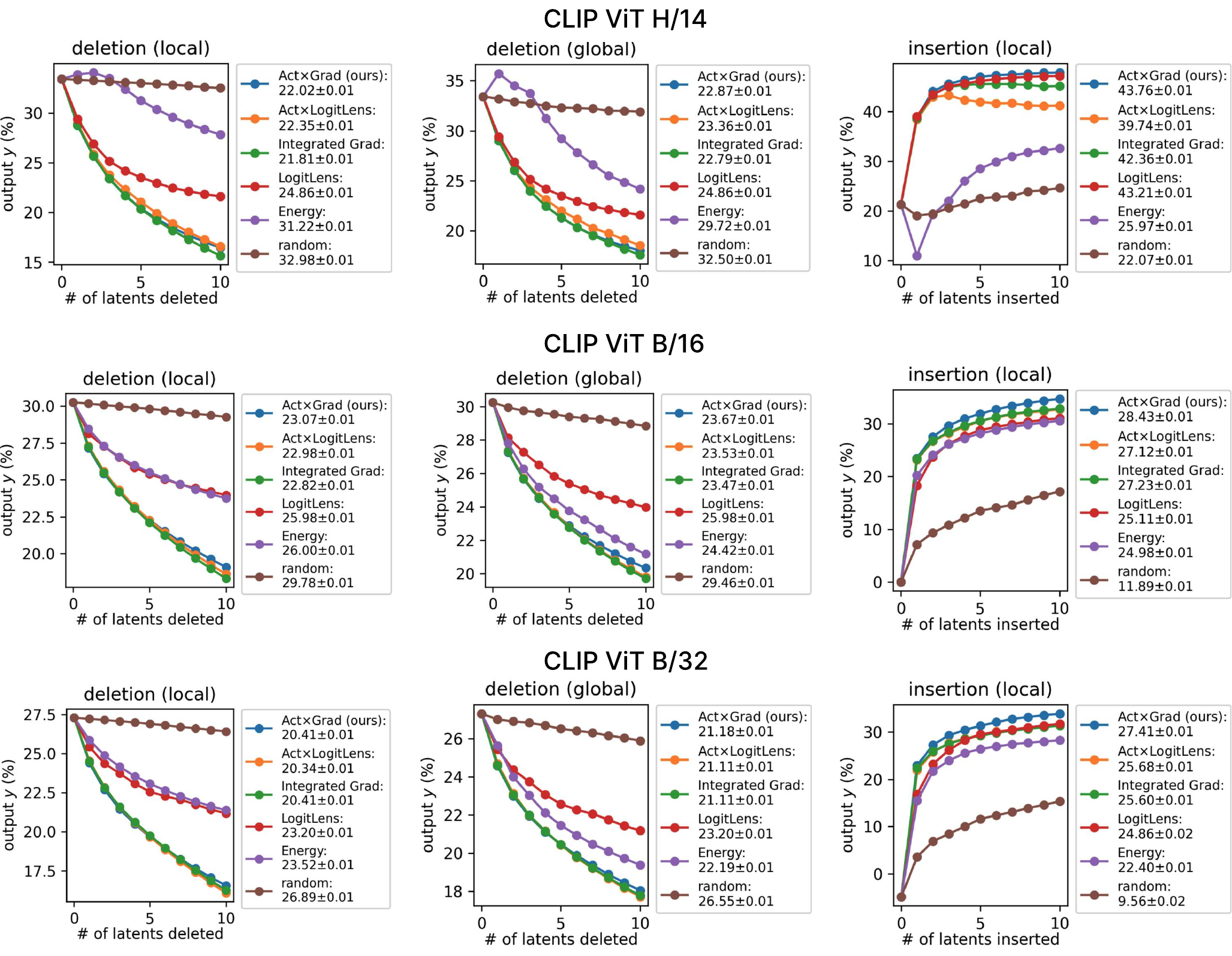}
    \caption{Evaluating Faithfulness of attributions of different attribution methods under three evaluation settings: local deletion (\emph{left}), {global deletion} (\emph{middle}), and {local insertion} (\emph{right}). Gradient-based approaches (e.g., Act$\times$Grad, Integrated Gradients) consistently outperform baselines, indicating higher faithfulness. Methods relying solely on activation magnitude or alignment (e.g., Energy, Logit Lens) show limited reliability, especially in early steps. Errors represent standard error of the mean across samples.}
    \label{app:fig:attributions}
\end{figure}
\begin{figure}
    \centering
    \includegraphics[width=1\linewidth]{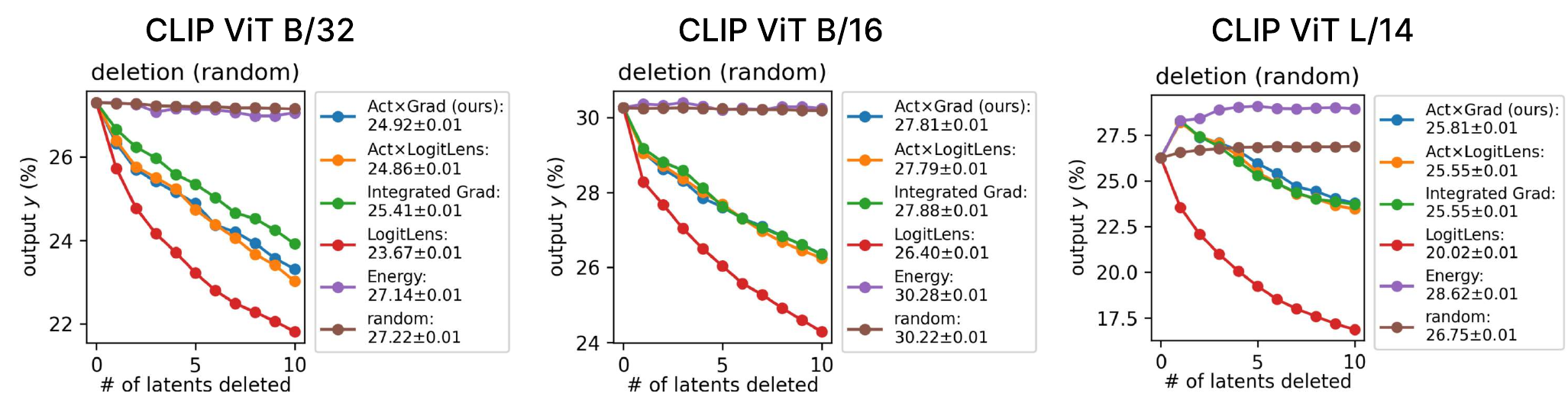}
    \caption{Evaluating Faithfulness of attributions of different attribution methods when attribution scores are estimated on random reference samples. All reference-sample dependent methods perform significantly worse in this setting, with Logit Lens resulting in the best scores. This underscores the need to chose meaningful reference samples. Errors represent standard error of the mean across samples.}
    \label{app:fig:attributions_random}
\end{figure}
The results are summarized in \cref{app:fig:attributions}. Across all models and settings, random attribution order consistently yields the worst performance. The energy-based attributions also shows limited faithfulness, and shows an interesting outlier behavior for ViT-H/14 and ViT-L/14 in the first latent deletion or insertion step. This suggests that highly activated components can exist that do not contribute to the model output and may even counteract it during deletion. This observation highlights that high activation alone does not imply relevance or alignment with the output.

While Logit Lens, which measures alignment with the output, performs better than energy-based attributions, it still falls short as it ignores component magnitude. The Act$\times$LogitLens approach, which combines activation magnitude with alignment, achieves performance comparable to Act$\times$Grad and Integrated Gradients for deletion. However, for insertion, it often underperforms. This discrepancy may be due to the fact that in early insertion steps, alignment (and component interaction) matters more than activation strength. Such interactions are better captured by the local correction term introduced in \cref{eq:methods:approximation}, which gradient-based methods implicitly account for.

Integrated Gradients generally outperforms other methods when a larger number of components are deleted. Across both insertion and deletion tasks, gradient-based methods show the best overall performance. It is worth noting, however, that Integrated Gradients is significantly more computationally expensive -- approximately $N$ times more than Act$\times$Grad, depending on the number $N$ of integration steps used.

Notably,
the choice of reference samples is crucial as demonstrated in \cref{app:fig:attributions_random},
where we show deletion experiment results when attributions are erstimated on randomly chosen reference samples (and not the actual class-specific reference samples).
Here, Logit Lens performs significantly better than all other baselines.

\section{Interpretability and diversity of SAE components}
\label{app:sec:sae_interpretability}
This section corresponds to \cref{sec:exp:sae} of the main manuscript, and provides more quantitative and qualitative results for SAE component interpretability analysis and concept diversity comparisons.

\subsection{Interpretability correlates with activation magnitude}

We expand on the analysis from \cref{sec:exp:sae}, examining how interpretability (measured as clarity using CLIP-Mobile-S2 and the 20 most activating samples on the ImageNet-1k train set) relates to the magnitude and frequency of component activations. Specifically, we consider three metrics:

\begin{enumerate}
    \item Mean activation on the top-5 most activating images per component,
    \item Average activation over the entire dataset,
    \item Firing rate, defined as the number of non-zero activations per component.
\end{enumerate}
\begin{figure}
    \centering
    \includegraphics[width=1\linewidth]{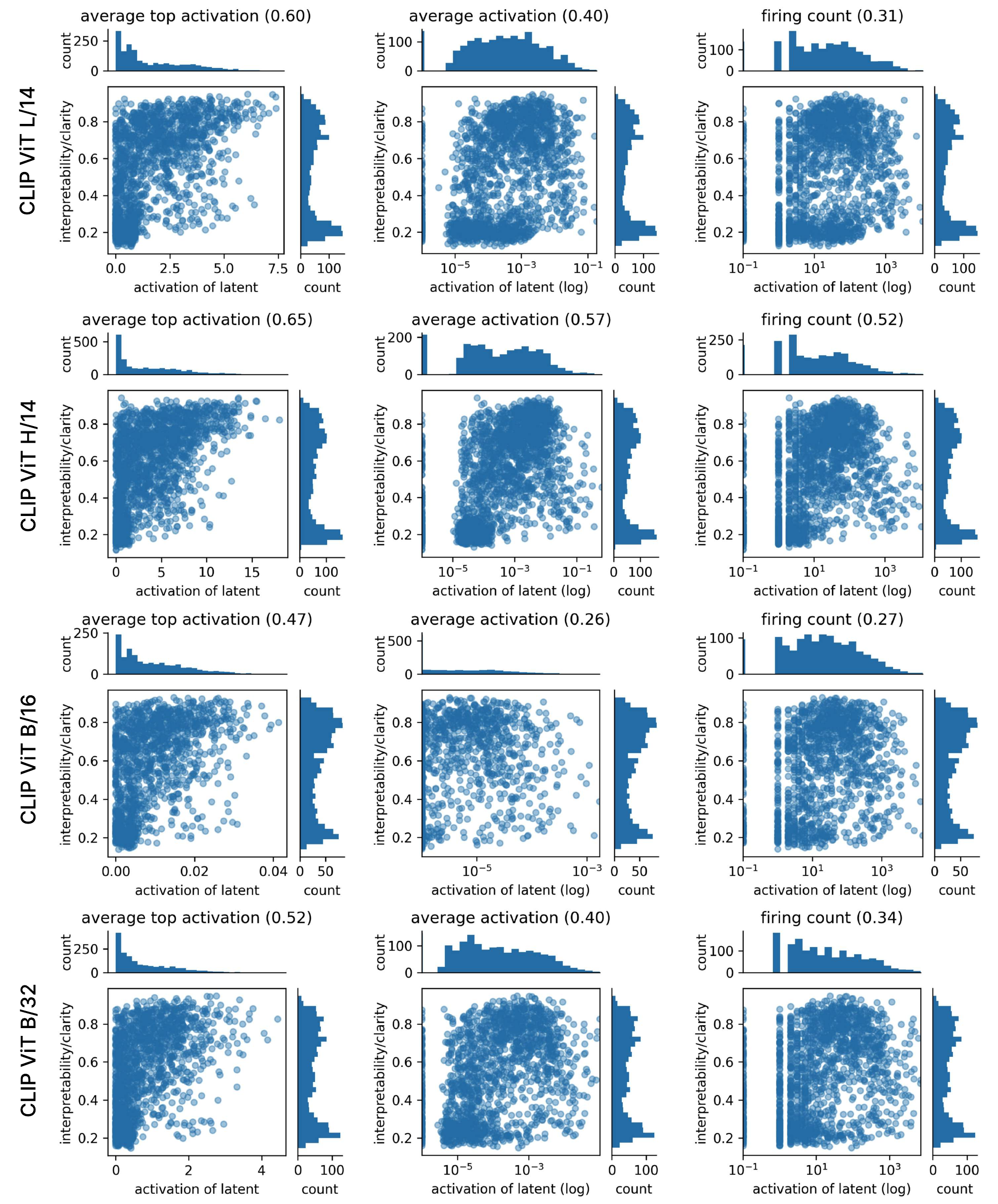}
    \caption{Correlation between interpretability (clarity) and component activation metrics across ImageNet-1k. We compare top-5 activation averages, overall mean activations, and firing rate (non-zero count). Top-5 activations show the strongest correlation with clarity. Average and count-based measures are shown on a logarithmic scale to highlight positive, though weaker, correlations.
    }
    \label{app:fig:sae_interpretability:correlation}
\end{figure}
All measurements are now computed on the ImageNet-1k \emph{test} set (results in the main manuscript were performed on the train set). Results for ViT-L/14, ViT-H/14, ViT-B/16, and ViT-B/32 models are shown in \cref{app:fig:sae_interpretability:correlation}.
We observe the following trends:
Top-5 activations exhibit the strongest correlation with interpretability (correlation coefficients are reported in the titles of each scatter plot). In contrast, average activation and firing rate show weaker correlations unless log-transformed, suggesting their raw distributions are highly skewed. Notably, the positive correlation between firing rate and clarity contradicts the common assumption that sparsity promotes interpretability, indicating that more frequently active components may also encode coherent and meaningful features.

\subsection{Finding unexpected components and concepts}
\begin{figure}
    \centering
    \includegraphics[width=1\linewidth]{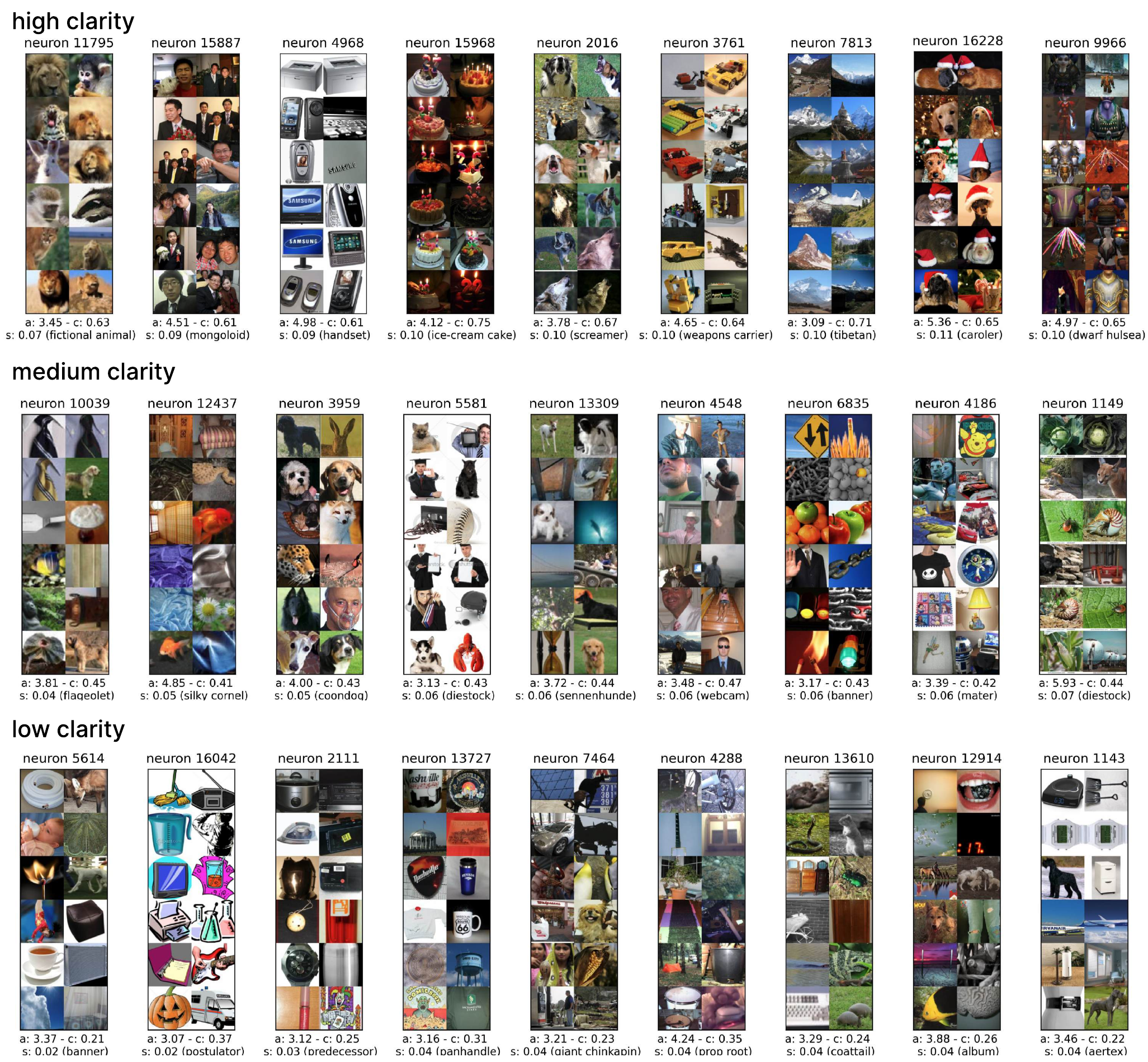}
    \caption{Examples of unexpected concepts found in CLIP ViT-L/14 for different levels of interpretability (clarity $c$). Components are grouped by interpretability clarity: high ($c \geq 0.6$), medium ($0.4 < c < 0.6$), and low ($c < 0.4$). For each component, we report the average activation $a$ across the top 20 activating samples and the textual alignment score $s$ with the best-matching label. 
    }
    \label{app:fig:sae_diversity_examples_1}
\end{figure}

\begin{figure}
    \centering
    \includegraphics[width=1\linewidth]{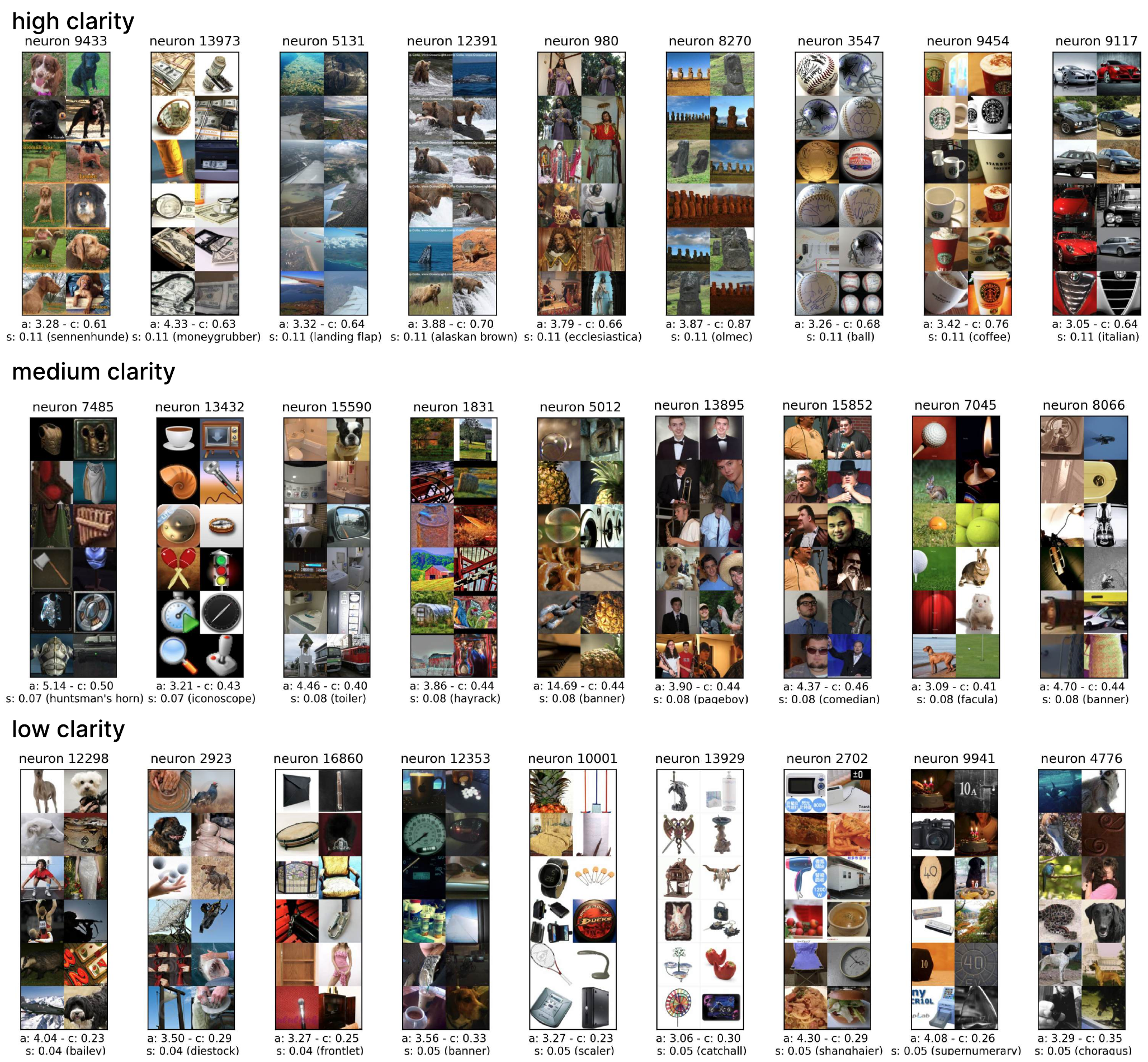}
    \caption{Examples of unexpected concepts found in CLIP ViT-L/14 for different levels of interpretability (clarity $c$). Components are grouped by interpretability clarity: high ($c \geq 0.6$), medium ($0.4 < c < 0.6$), and low ($c < 0.4$). For each component, we report the average activation $a$ across the top 20 activating samples and the textual alignment score $s$ with the best-matching label. }
    \label{app:fig:sae_diversity_examples_2}
\end{figure}
We present additional examples of latent components discovered using the approach described in~\cref{sec:methods:attribution_analysis}a. 
Specifically, we filter components with an average activation magnitude greater than $3.0$ over their top-20 activating samples, 
and with low alignment to ImageNet-21k class labels (as defined in~\cref{eq:background:labeling}); 
see~\cref{app:fig:sae_diversity_examples_1,app:fig:sae_diversity_examples_2} for examples.
Here, activation is used as a proxy for relevance. 
While activation alone is not always indicative of importance for a specific output, it is nevertheless informative in this setting, 
as it provides a model-internal signal independent of a particular prompt. 
According to~\cref{eq:methods:approximation}, activation contributes directly to the output, 
thus high-activation components may influence predictions across a wide range of prompts.

These components are further grouped by their interpretability levels (clarity). 
We observe diverse and unexpected concepts, including textual artifacts such as image captions 
(e.g., component $\#1149$ in~\cref{app:fig:sae_diversity_examples_1}), watermarks (component $\#5581$ in~\cref{app:fig:sae_diversity_examples_1}), 
and typographic elements like the letter/number ``O/0'' ($\#9941$ in~\cref{app:fig:sae_diversity_examples_2}). 
Additionally, some components appear to encode brand-specific information, 
such as car logos (component $\#9117$ in~\cref{app:fig:sae_diversity_examples_2}) or electronic branding (component $\#4968$ in~\cref{app:fig:sae_diversity_examples_1}).

\section{Robustness analysis of CLIP}
\label{app:sec:robustness}
This section provides more details for experiments of \cref{sec:exp:robustness}, where we benchmark CLIP's robustness to spurious correlations, polysemous prompts, compound nouns, and visual typography.
We first applied approach \cref{sec:methods:attribution_analysis}b to the CLIP ViT-B/16 model to find outlier components, which were relevant on high-confidence samples, but were not expected to be relevant.

\subsection{Finding failure modes}
We conduct our analysis on a 20\% subset of the ImageNet-1k training set, obtained by sampling every fifth image.
To identify potentially misleading SAE components, we first estimate their relevance distributions on class samples (approximately 200 per class), computing both mean and standard deviation of relevance scores.
Relevance scores are computed using \cref{eq:methods:relevance} by explaining the cosine similarity between visual embedding and textual embedding for the full class label.
Next, we filter for samples (not belonging to the probed-for class) with high model output -- specifically, those exceeding the class-specific mean output minus 1.5 standard deviations.
For these high-confidence samples, we compute relevance scores and flag components with a resulting $z$-score above 3.0 and at least 10 non-zero activations as potential failure cases.
These components are manually reviewed and incorporated into targeted test sets.

\begin{figure}
    \centering
    \includegraphics[width=1\linewidth]{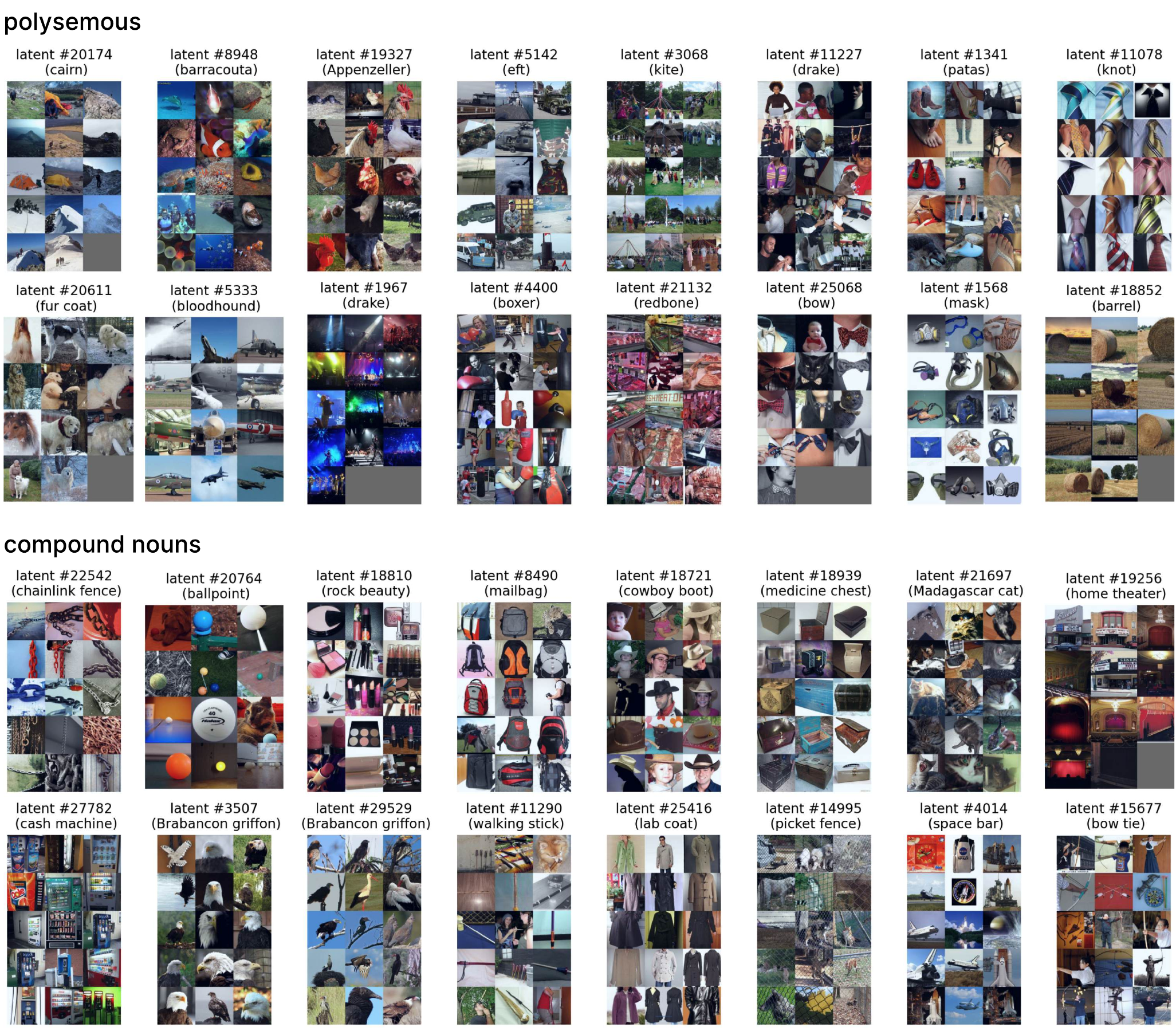}
    \caption{Examples of sample sets where text-image probing led to false positives, corresponding to polysemous words or compound nouns. Each set contains high-confidence, highly activating samples for a flagged SAE component. The latent ID and the prompt class name are shown for each set.}
    \label{app:fig:failure_cases:examples_1}
\end{figure}
\begin{figure}
    \centering
    \includegraphics[width=1\linewidth]{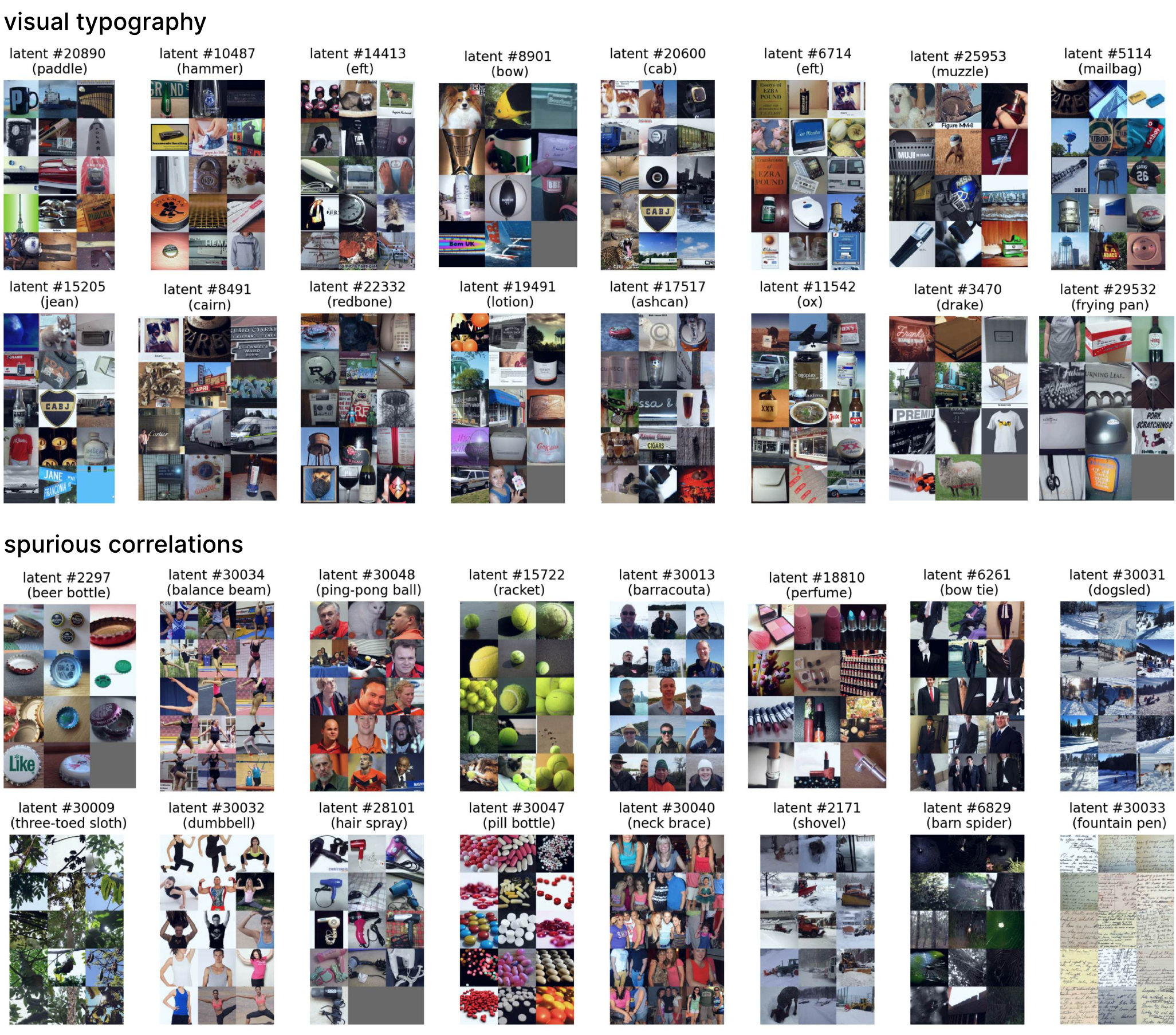}
    \caption{Examples of sample sets where text-image probing led to false positives, corresponding to visual typography or spurious correlations. Each set contains high-confidence, highly activating samples for a flagged SAE component. The latent ID and the prompt class name are shown for each set. Spurious correlations for components with ID above 30,000 correspond to spurious correlations found by \cite{neuhaus2023spurious}.}
    \label{app:fig:failure_cases:examples_2}
\end{figure}

In total, we identify 76 polysemous failure modes, 78 compound noun cases, 21 related to visual typography, and 57 involving spurious correlations. 
Representative examples from each category are shown: polysemous and compound noun cases in \cref{app:fig:failure_cases:examples_1}, and visual typography and spurious correlations in \cref{app:fig:failure_cases:examples_2}, with 16 examples per set.

\begin{figure}
    \centering
    \includegraphics[width=1\linewidth]{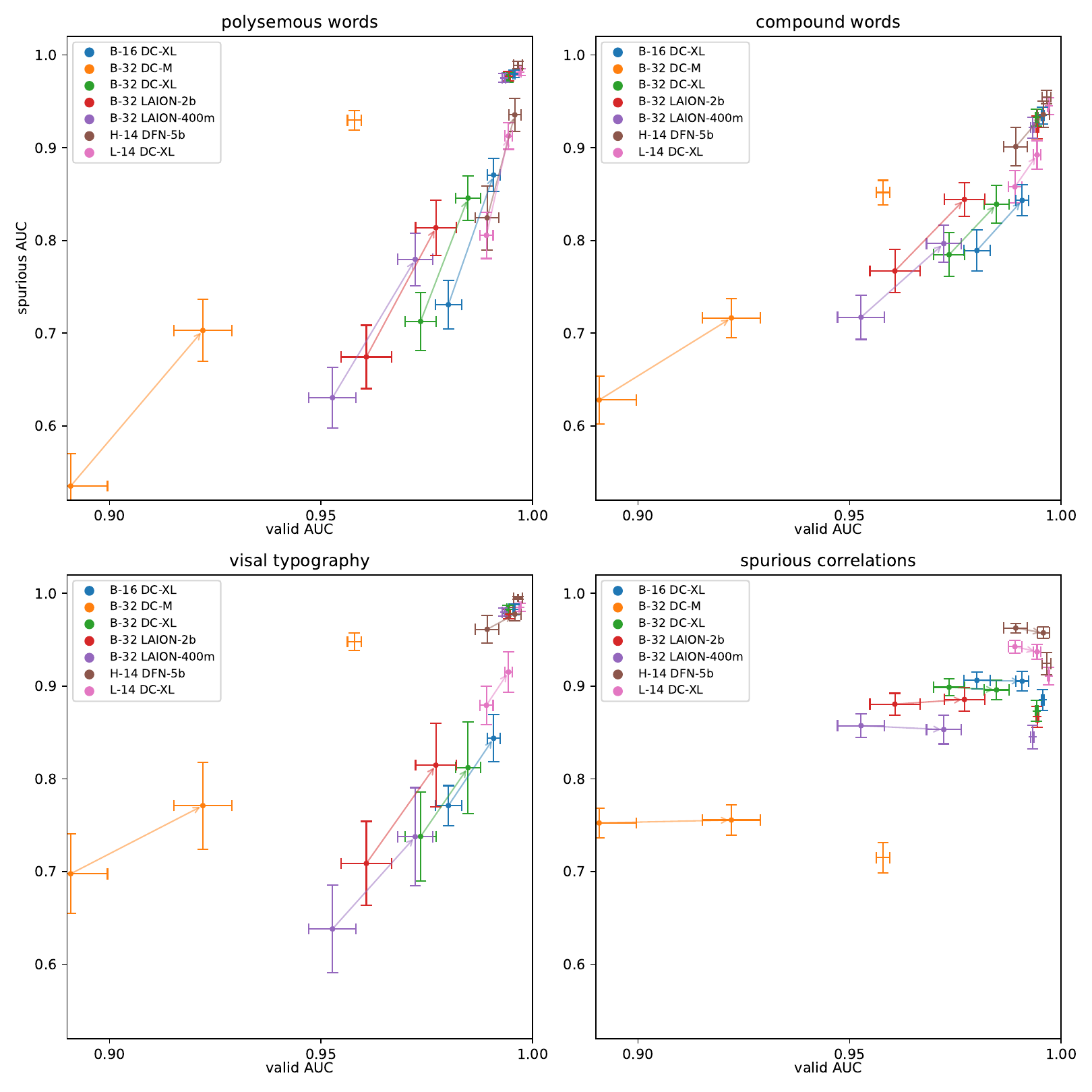}
    \caption{A detailed plot variant of \cref{fig:experiments:benchmark} for the analysis of failure modes in text-image probing. We evaluate CLIP’s robustness to textual ambiguities and dataset artifacts across multiple model variants. Failure cases include polysemous words, compound nouns, typography in images, and spurious correlations. For each scenario, we assess the separability of true images from found distractors (``spurious AUC'') and from unrelated ImageNet classes (``valid AUC'').
    We compare performance using both vanilla and enriched text prompts (e.g., via templating and detailed descriptions), and include linear classifiers trained on image embeddings as a baseline.}
    \label{app:fig:failure_cases:extended_plot}
\end{figure}

\subsection{Evaluating failure modes}
We provide a detailed plot of \cref{fig:experiments:benchmark} in \cref{app:fig:failure_cases:extended_plot}, where we also include the ViT-B/32 model trained on the DataComp-M dataset. We also display \gls{sem} values for each measurement point. The \gls{sem} is computed by assuming that the measured AUC scores are normally distributed.

We evaluate text-image probing using three methods:
(1) the short ImageNet-1k class name (excluding text after a comma),
(2) the full class name with the mean textual embedding from three templates (``\texttt{object}'', ``an image of a \texttt{object}'', and ``a \texttt{object}''),
and (3) linear regression classifiers trained on image embeddings from the test set.
For identified failure modes, we assess the ability to distinguish true class samples from spurious ones by computing AUROC scores on the model outputs (``spurious AUC'').
Additionally, we evaluate how well true class samples can be separated from all other non-class samples in the training set (``valid AUC'').
Ideally, both spurious and valid AUC scores are high.

\begin{table}[t]
\centering
\caption{Investigating how better textual descriptions and templating effects failure modes detected in CLIP. We report the difference in spurious AUC scores (mean with \gls{sem}) across different failure case types and prompting variants}
\begin{tabular}{lccc}
\toprule
\textbf{Failure Case} & \textbf{Templating} & \textbf{Extended Description} & \textbf{Both} \\
\midrule
Polysemous words & 0.020 ± 0.032 & 0.135 ± 0.028 & 0.151 ± 0.028 \\
Compound nouns   & -0.001 ± 0.022 & 0.060 ± 0.020 & 0.055 ± 0.021 \\
Visual typography       & 0.063 ± 0.033 & 0.068 ± 0.038 & 0.117 ± 0.033 \\
Spurious correlations   & -0.002 ± 0.011 & -0.001 ± 0.011 & -0.005 ± 0.012 \\
\bottomrule
\label{app:tab:failure_case}
\end{tabular}
\end{table}

The influence of class name length and prompt templating on performance is shown in \cref{app:tab:failure_case}.
We observe that extended class descriptions notably improve performance for polysemous terms and compound nouns.
Prompt templating yields a modest benefit only for visual typography cases, likely because longer prompts dilute the impact of short typographic artifacts by increasing the overall character count.
In contrast, spurious correlations appear largely unaffected by either class name length or templating.

\section{Medical case study: melanoma detection}
\label{app:sec:medical_case_study}
This section provides additional methodological details for the experiments described in \cref{sec:exp:medical}, where we evaluate WhyLesion-CLIP on the ISIC 2019 dataset.

We construct a binary classification task by randomly selecting 80\,\% of melanoma samples (corresponding to 3,629 images) and 4,000 non-melanoma images to train linear classifiers (logistic regression models). These classifiers operate on the CLIP-derived visual embeddings.

\paragraph{Image-level red hue augmentation}
To simulate the influence of red hue, we manipulate the red channel of input images. Let \texttt{red}~$\in[0, 1]$ denote the red color channel of an image. We define the augmented red channel as:
\begin{equation}
    \texttt{red}'= \text{max}[\text{min}(\texttt{red} (1-\delta), 1), 0]~,
\end{equation}
where pixel values are clamped to remain within $[0,1]$. We apply $\delta = -1.0$ to increase red hue and $\delta = 0.2$ to reduce it.

\begin{figure}
    \centering
    \includegraphics[width=1\linewidth]{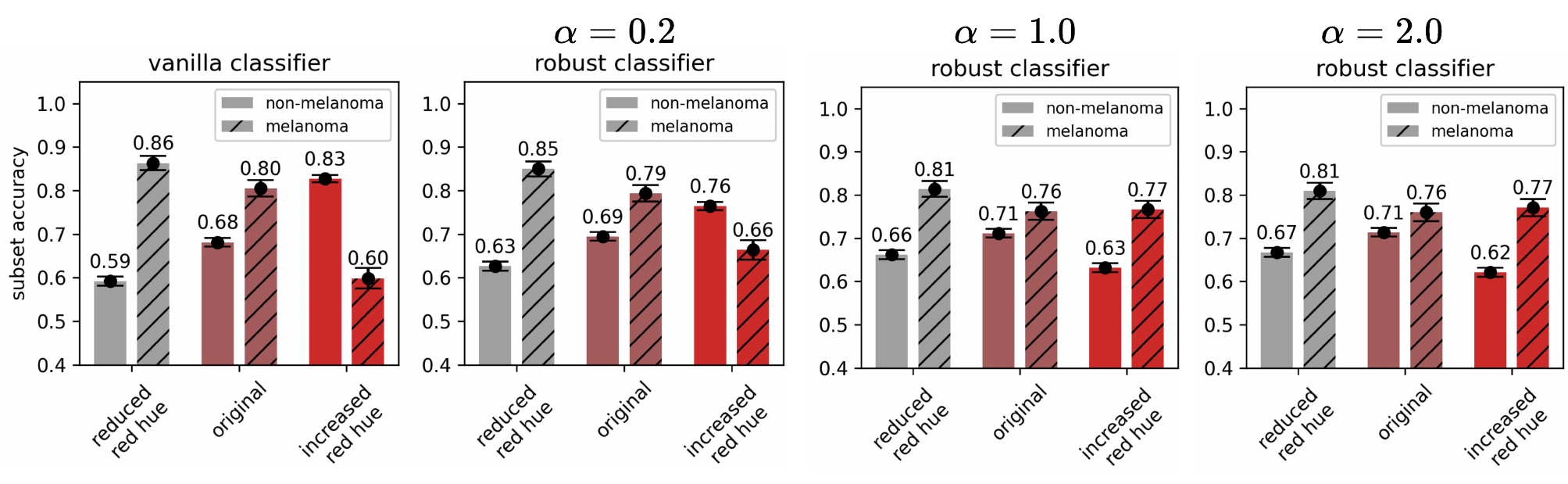}
    \caption{Robustness analysis of linear classifiers trained on WhyLesion-CLIP visual embeddings for melanoma detection.
In addition to training standard (non-augmented) classifiers, we train robust models using latent space augmentations of varying strengths, controlled by the parameter $\alpha$. Results show that increasing $\alpha$ improves robustness, as performance degrades less under augmentation, indicating enhanced invariance to spurious red hue correlations.}
    \label{app:fig:medical_case_study:other_hyperparameters}
\end{figure}
\paragraph{Latent-level red hue augmentation}
To perturb embeddings in a semantically meaningful direction, we first estimate a ``red hue'' direction in latent space. We identify this direction by comparing two subsets of images using SAE component $\#2703$, which is highly sensitive to red tones. Specifically,
(1) the low-red set includes non-melanoma samples where $a_{2703} < 1.0$,
and (2) high-red set includes non-melanoma samples where $a_{2703} > 2.5$.
For reference, the mean activation of component $\#2703$ is $1.17$ with standard deviation $0.65$.

We compute the red-direction vector $\boldsymbol{u}_\text{red}$ in latent space as the mean embedding difference: 
\begin{equation}
    \boldsymbol{u}_\text{red} = \bar{\boldsymbol{x}}_+ - \bar{\boldsymbol{x}}_-~,
\end{equation}
where $\bar{\boldsymbol{x}}_+$ and $\bar{\boldsymbol{x}}_-$ denote the average latent embeddings of the high-red and low-red sets, respectively.

To apply this augmentation, we sample a scalar $p \sim \mathcal{U}(0, 1)$ and modify the original latent embedding $\boldsymbol{x}$ as:
\begin{equation}
    \boldsymbol{x}' = \boldsymbol{x} + \alpha(p-0.5)\boldsymbol{u}_\text{red},
\end{equation}
where $\alpha$ is a scaling parameter. 
In the main experiment reported in \cref{sec:exp:medical}, we set $\alpha = 0.5$.
Results for alternative values of $\alpha$ are presented in \cref{app:fig:medical_case_study:other_hyperparameters}.
The plot shows,
that the larger $\alpha$, the less effected is the performance (subset accuracy) by the input augmentation.

\end{document}